%% file: acl_main.tex
\documentclass[11pt]{article}

\usepackage[final]{acl}

\usepackage{times}
\usepackage{latexsym}

\usepackage[T1]{fontenc}

\usepackage[utf8]{inputenc}

\usepackage{microtype}

\usepackage{inconsolata}

\usepackage{graphicx}

\usepackage{booktabs}       
\usepackage{amsfonts}       
\usepackage{nicefrac}       
\usepackage{microtype}      
\usepackage{xcolor}         
\usepackage{xspace}
\usepackage{graphicx} 
\usepackage{amsmath} 
\usepackage{bm}
\usepackage{amssymb}
\usepackage{algorithm}
\usepackage{setspace}
\usepackage{algpseudocode} 
\usepackage{graphicx}
\usepackage{amsmath}
\usepackage{arydshln}
\usepackage{amssymb}
\usepackage{booktabs}
\usepackage{bbm}
\usepackage[utf8]{inputenc} 
\usepackage{bbm, dsfont}
\usepackage{mathtools}
\usepackage{amsthm}
\usepackage[normalem]{ulem}
\usepackage{subcaption}
\usepackage{colortbl}
\usepackage{multirow}
\usepackage{pifont}
\usepackage{cutwin}
\usepackage{caption}
\usepackage{wrapfig}
\usepackage{microtype}
\usepackage{graphicx}
\usepackage{bm}
\usepackage{eqparbox}
\usepackage{makecell}
\usepackage{threeparttable}
\usepackage{adjustbox}
\usepackage{xspace}
\usepackage{titletoc}
\usepackage{tocloft}
\usepackage{enumitem}
\input{macros}

\newcommand{\eg}{\emph{e.g.}\xspace}
\newcommand{\ie}{\emph{i.e.}\xspace}

\newcommand{\ours}[1]{\textsc{VideoRepair}}
\newcommand{\dsgours}[1]{VSQ}    

\definecolor{red}{RGB}{180,0,0} 
\definecolor{bblue}{RGB}{0,0,0} 
\definecolor{gg}{RGB}{100,100,100}

\title{Self-Correcting 
Text-to-Video Generation \\
with 
Misalignment Detection and Localized Refinement}

\author{
Daeun Lee$^{1}$ \quad Jaehong Yoon$^{2}$ \quad Jaemin Cho$^{3,4}$ \quad Mohit Bansal$^{1}$\\ 
$^{1}$UNC Chapel Hill\quad
$^{2}$NTU Singapore\quad 
$^{3}$Allen Institute for AI\quad 
$^{4}$Johns Hopkins University
\\ 
\url{https://video-repair.github.io/}
}

\definecolor{step1}{HTML}{E7F2DC}
\definecolor{step2}{HTML}{FBF5DC}
\definecolor{step3}{HTML}{FBE4E7}

\begin{document}


\twocolumn[{%
    \renewcommand\twocolumn[1][]{#1}%
    \vspace{-2em}
    \maketitle
    \begin{center}
        \centering
        \vspace{-0.2in}
            \includegraphics[width=0.95\linewidth]{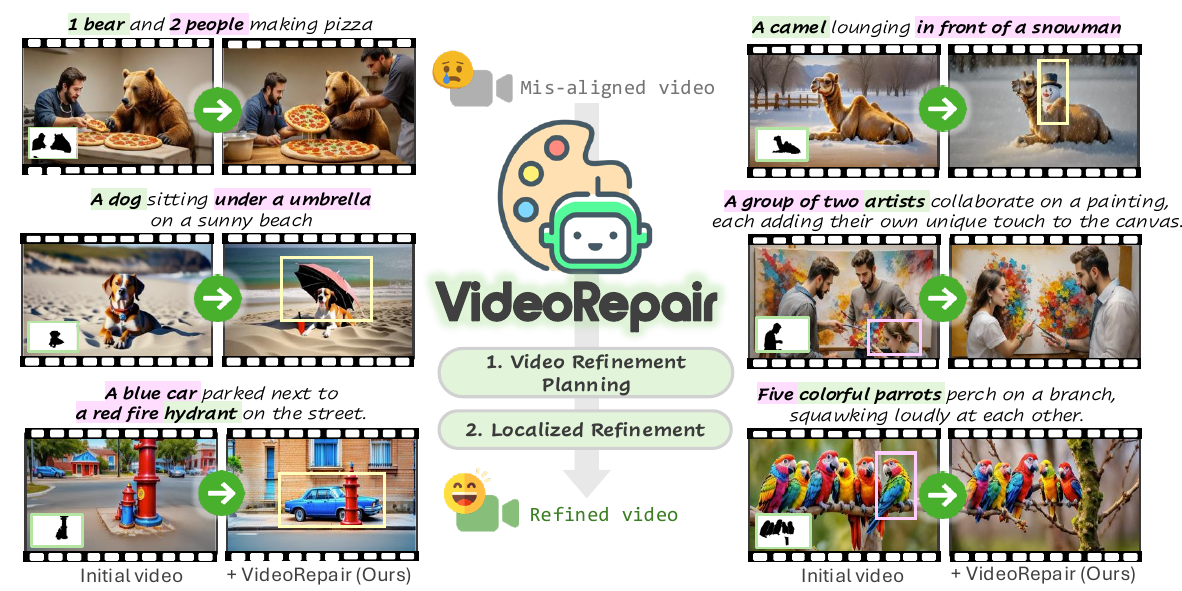}
        \captionof{figure}{
        \textbf{
        \ours{}} is a model-agnostic, training-free refinement framework for improving alignments in text-to-video generation. 
        Given an initial video from a text-to-video generation model, \ours{} refines the generated video in three stages:
    (1) misalignment detection (\cref{sec:subsec:video_evaluation}),
    (2) refinement planning (\cref{sec:subsec:planning}) and 
    (3) localized refinement (\cref{sec:subsec:localized_refinement}). 
        The black-white mask in the bottom left of each example indicates the localized refinement plan (black: regions to preserve / white: regions to refine).
        }
        \label{fig:teaser_twocolumn}
        \vspace{5pt}
    \end{center}
}]

\input{sec/01_abstract}

\input{sec/02_intro}

\input{sec/04_method}

\input{sec/05_experiments}

\input{sec/03_related}

\input{sec/06_conclusion}

\bibliography{custom}
\appendix

\input{sec/Appedix}

\end{document}

%% file: macros.tex
\definecolor{darkgreen}{rgb}{0,0.5,0}
\definecolor{azureblue}{rgb}{0,0.5,1}
\definecolor{darkgreen}{rgb}{1,0,0}
\definecolor{color1}{HTML}{006EB8}
\definecolor{color2}{HTML}{009B55}
\definecolor{color3}{HTML}{924DBF}

\usepackage{pifont}

\hypersetup{
    colorlinks=true,
    linkcolor=magenta,
    filecolor=magenta,  
    urlcolor=magenta,
    citecolor=color3,
    pdftitle={Overleaf Example},
    pdfpagemode=FullScreen,
    }

\usepackage[capitalize]{cleveref}
\crefname{section}{Sec.}{Secs.}
\Crefname{section}{Sec.}{Secs.}
\Crefname{table}{Tab.}{Tabs.}
\crefname{table}{Tab.}{Tabs.}
\Crefname{figure}{Fig.}{Figs.}
\crefname{figure}{Fig.}{Figs.}
\crefname{appendix}{Sec.}{Secs.}
\Crefname{appendix}{Tab.}{Tabs.}

\usepackage{color,colortbl}
\definecolor{darkgreen}{rgb}{0,0.5,0}

\definecolor{darkgreen}{rgb}{1,0,0}

\definecolor{azureblue}{rgb}{0,0.5,1}

%% file: sec/01_abstract.tex
\begin{abstract}

{Recent text-to-video (T2V) diffusion models have made remarkable progress in generating high-quality videos. 
However, they often struggle to align with complex text prompts, particularly when multiple objects, attributes, or spatial relations are specified. 
We introduce \textbf{\ours{}}, the first self-correcting, training-free, and model-agnostic video refinement framework that automatically detects fine-grained text–video misalignments and performs targeted, localized corrections. 
Our key insight is that even misaligned videos usually contain correctly generated regions that should be preserved rather than regenerated. 
Building on this observation, \ours{} proposes a novel region-preserving refinement strategy with three stages: 
(i) \textit{misalignment detection}, where MLLM-based evaluation with automatically generated evaluation questions identifies misaligned regions; 
(ii) \textit{refinement planning}, which preserves correctly generated entities, segments their regions across frames, and constructs targeted prompts for misaligned areas; and 
(iii) \textit{localized refinement}, which selectively regenerates problematic regions while preserving faithful content through joint optimization of preserved and newly generated areas. 
On two benchmarks, EvalCrafter and T2V-CompBench with four recent T2V backbones, \ours{} achieves substantial improvements over recent baselines across diverse alignment metrics. 
Comprehensive ablations further demonstrate the efficiency, robustness, and interpretability of our framework.
}

\end{abstract}

%% file: sec/02_intro.tex
\section{Introduction}
\label{sec:intro}

\begin{figure*}[t]
\centering

\begin{minipage}[t]{0.33\textwidth}
    \centering
    \includegraphics[width=\textwidth]{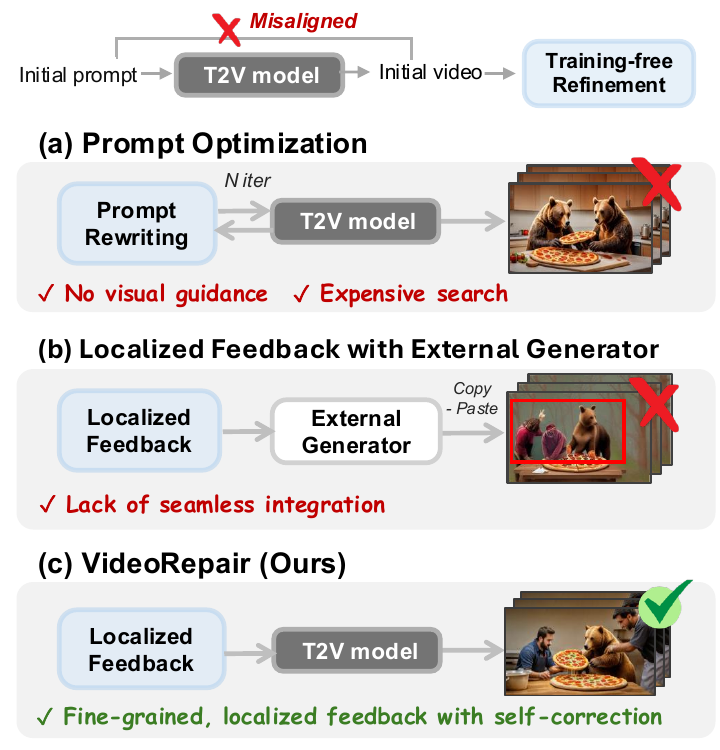}
    \vspace{-0.2in}
    \captionof{figure}{
    \textbf{Comparison with other baselines.}
    \ours{} provides localized feedback with self-correcting. 
    }
    \label{fig:comparison}
\end{minipage}
\hfill
\begin{minipage}[t]{0.63\textwidth}
    \centering
    \includegraphics[width=\textwidth]{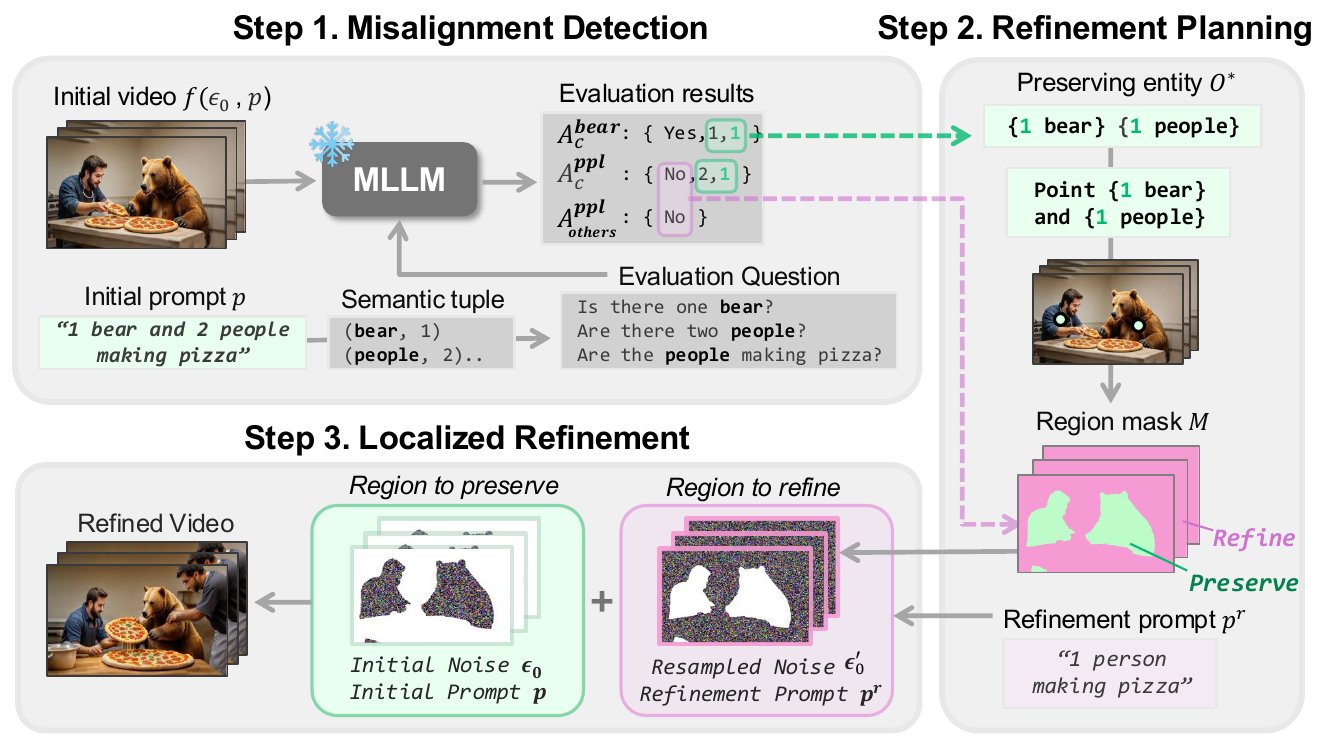}
    \vspace{-0.2in}
    \captionof{figure}{
    \textbf{Illustration of \ours{}.}
    \ours{} refines the generated video in three stages:
    (1) misalignment detection (\cref{sec:subsec:video_evaluation}),
    (2) refinement planning (\cref{sec:subsec:planning}) and 
    (3) localized refinement (\cref{sec:subsec:localized_refinement}).
    }
    \label{fig:method}
\end{minipage}

\vspace{-0.in}
\end{figure*}

Recent text-to-video (T2V) diffusion models~\citep{ho2022video,singer2022make,esser2023structure,blattmann2023stable,khachatryan2023text2video,wang2023modelscope,yang2024cogvideox,wan2025wan} have achieved impressive photorealism and versatility across diverse domains.
Despite these advances, current models often struggle to faithfully follow input text prompts, especially when the prompt specifies multiple objects and attributes. Typical errors include generating the wrong number of objects, mismatched attribute bindings, or distorted regions.

To mitigate these issues, recent works~\citep{vico,tian2024videotetris,qu2025ttom} propose compositional T2V techniques that improve text–video alignment. While these methods enhance compositionality, they lack explicit feedback mechanisms to detect and correct misalignments, limiting their adaptability and interpretability in real-world scenarios. 
In parallel, several image-based studies~\citep{opt2i,sld,chen2025t2i,ji2025prompt,xiang2025promptsculptor} have introduced training-free frameworks that refine outputs using guidance from LLMs or MLLMs. 
However, as shown in \cref{fig:comparison}, these approaches are computationally expensive, dependent on external generators, or prone to visual inconsistencies. 

To address these challenges, we introduce \ours{}, the first self-correcting framework for text-to-video generation that is compatible with any diffusion-based T2V backbone in a training-free manner. 
Our key insight is that even when generated videos contain misaligned or distorted objects, certain key elements are often accurately generated in specific regions.
Similar to how humans revise creative work by fixing only the errors while keeping what is correct, \ours{} preserves accurately generated regions and selectively refines only the problematic ones.
This region-preserving strategy leverages diffusion models’ natural ability to regenerate content from noise while avoiding unnecessary changes to faithful areas.
Moreover, detecting correctly generated content via grounding and segmentation is substantially easier than exhaustively enumerating all possible distortions, making our approach both reliable and efficient.

Building on this intuition, \ours{} implements region-preserving refinement through three mutually reinforcing processes: \textit{misalignment detection}, \textit{refinement planning}, and \textit{localized refinement} as illustrated in \cref{fig:method}.
Unlike prior refinement frameworks that operate on the entire video indiscriminately, \ours{} follows a self-correcting, region-preserving paradigm: it distinguishes correctly generated regions from misaligned ones and regenerates only the latter.
This transforms evaluation feedback into actionable generative guidance, allowing precise corrections without discarding high-quality content and establishing a new paradigm for efficient, interpretable video refinement.
Specifically, spatio-temporal evaluation questions derived from the prompt expose fine-grained errors; these signals guide the selection of entities to preserve and the construction of a targeted refinement prompt; and localized regeneration is then harmonized with preserved regions to yield perceptually seamless videos.

We validate \ours{} on two challenging benchmarks, EvalCrafter~\citep{liu2023evalcrafter} and T2V-CompBench~\citep{sun2024t2v}, which cover diverse prompt categories including object counts, spatial relations, and global scene attributes. 
Empirically, \ours{} substantially outperforms existing refinement methods across a wide range of compositional prompts, while preserving global quality aspects such as visual fidelity, motion smoothness, and temporal consistency. 
We further provide detailed ablations on each component, error accumulation, inference latency, and robustness to different MLLM replacements, underscoring the generality and reliability of our framework.

Our key contributions are as follows:
\vspace{-0.05in}
\begin{itemize}[leftmargin=1em]
\item We present the first self-correcting, training-free framework for text-to-video generation, compatible with diffusion-based T2V backbones, that detects misalignments via MLLM-based evaluation and plans targeted refinements.
\item We introduce a region-preserving refinement strategy that transforms evaluation feedback into actionable guidance, preserving correct regions while selectively regenerating misaligned ones, offering both effective correction and interpretable feedback.
\item We show that \ours{} consistently improves text–video alignment across diverse models and benchmarks, while maintaining fidelity, temporal coherence, and motion quality, outperforming all prior training-free refinement approaches.
\end{itemize}

%% file: sec/04_method.tex
\section{\ours{}} 

\label{sec:method}

We introduce \ours{}, the first \textit{training-free, model-agnostic
self-correcting} framework for text-to-video generation.
Unlike prior refinement approaches that either use only prompt optimization~\citep{opt2i} or rely on external generative models~\citep{sld}, our approach
follows a new principle: 
\textit{preserve the correct region, selectively repair where it is wrong}. 
This principle distinguishes \ours{} from generic mask-based
inpainting or editing: the preserved regions are determined not by manual
masks or heuristic rules, but by an automatic video evaluation
and planning process that identifies semantically aligned objects
directly from the input prompt and the generated video.
By tightly coupling evaluation, planning, and refinement within
the same T2V backbone, \ours{} enables localized regeneration that improves compositional fidelity while maintaining temporal consistency and visual realism.

\textbf{Problem statement.} 
Our goal is to improve text-video alignment using a pre-trained T2V diffusion model $f(\cdot)$, without requiring any additional fine-tuning. 
Given a text prompt $p$ and initial noise $\epsilon_0 \sim \mathcal{N}(0, \boldsymbol{I})$, 
we generate an initial video 
$V_0 = f(p, \epsilon_0)$, where $V_0 \in \mathbb{R}^{3 \times H \times W \times T}$ and $H$, $W$, and $T$ denote the height, width, and number of frames, respectively. 
If $V_0$ exhibits misaligned or inaccurate content, we evaluate it using a set of questions derived from the prompt $p$ and construct a refinement plan. 
Then, we perform localized self-refinement with the same T2V model $f(\cdot)$, producing a refined video $V_1$ that better aligns with the original prompt. 

\subsection{\colorbox{step1}{Misalignment Detection} }\label{sec:subsec:video_evaluation}
\paragraph{Generate video evaluation questions.} 
To diagnose the initial video $V_0$, we generate video evaluation questions from $p$. 
Unlike prior question-based evaluations in the image domain~\citep{hu2023tifa,dsg}, these questions provide \emph{spatio-temporal feedback signals} that directly guide refinement planning. 
It goes beyond simple object-existence checks by explicitly capturing 
\emph{counts, attributes, spatio-temporal relations, actions, and scene-level global properties}, all of which are critical for faithful video-text alignment. 
Given a prompt $p$, we first extract a semantic tuple $\mathcal{T}$, 
a structured representation of entities, attributes, relationships, and actions relevant to the video. 
Using this as guidance, we employ in-context learning with an LLM to generate a set of evaluation questions $Q$. 
The resulting set $Q$ is divided into two disjoint subsets: 
$Q_c$ (questions focused on object counting) and $Q_{\text{others}}$ (questions covering all other aspects, such as action, attributes, and scene-level global properties), reflecting the distinct nature of count-based reasoning versus semantic understanding. 
To better support fine-grained counting, 
we let the LLM generate count-specific questions for individual objects 
(e.g., ``Is there \textit{one} bear?'') rather than merely verifying object existence 
(e.g., ``Is there a bear?''). 
Our ablation study (see \Cref{tab:ablation_components}) demonstrates that 
these evaluation questions provide more effective refinement guidance 
compared to existing question-based evaluation methods. 
Additional details are provided in the Appendix~\cref{sec:appendix-qa-detail}.

\paragraph{Answering to identify video errors.} 
We now evaluate $V_0$ to determine which region requires refinement, as illustrated in \cref{fig:method} (top right).
Given the entity set $O$ (\ie, object or scene element) from $\mathcal{T}$,
we group $Q^o = \{Q^o_c, Q^o_{\text{others}}\} \subset Q$ as the subset of questions that contain the name of $O$
Note that this entity captures not only localized object discrepancies but also global misalignments between $p$ and $V_0$
To this end, we employ an MLLM to answer each predefined question set $Q^o$ with binary judgments.
For count-related questions $Q_c^o$, we prompt the model to output both a binary decision and an estimated object count, resulting in a triplet
$A_c^o = \{b_c^o, n_p^o, n_v^o\}$,
where $n_p^o$ and $n_v^o$ denote the number of instances of object $o$ in the prompt $p$ and the video $V_0$, respectively.
The binary answer $b_c^o$ is set to 1 if $n_p^o = n_v^o$, and 0 otherwise.
For example, in \cref{fig:method}, the question “Is there one bear?” results in $b^{\text{bear}}_c = 1$ when both the prompt and the video indicate a single bear (i.e., $n_p^{\text{bear}} = n_v^{\text{bear}} = 1$).
For other type of questions $Q^o_{\text{others}}$, we prompt the model to return only a binary response $A^o_{\text{others}} = \{b_{\text{others}}^o\}$, where $b_{\text{others}}^o = 1$ indicates alignment between the element in $V_0$ and $p$, and $b_{\text{others}}^o = 0$ otherwise. 
If an entity disappears or becomes distorted across frames, we also regard it as a misalignment case.
We aggregate binary evaluation results into a video-level accuracy score in the range $[0,1]$. 
If the score is $1.0$, we terminate the process early, as the initial video is already fully correct. 
If the score is $0.0$, we instead re-generate the video with a new random seed to avoid uninformative outputs.

\subsection{\colorbox{step2}{Refinement Planning}}\label{sec:subsec:planning}

\paragraph{Identifying visual content to retain.} 
As mentioned earlier, \ours{} aims to retain accurately generated regions in the initial video while correcting only the mis-generated ones to ensure improved text-video alignment. 
To this end, we first identify the key entity $O^*$ and determine the number $N^*$ of its instances to be preserved. 
To select which entity should be retained, we prompt the MLLM with question--answer pairs and $V_0$ as input, allowing it to identify correctly generated entities to preserve.
For countable entities, the number of preserved instances $N^*$ is determined from the triplet $A^{o^*}_c=\{b^{o^*}_c, n^{o^*}_p, n^{o^*}_v\}$ as
\begin{equation}
N^* =
\begin{cases}
n_p^{o^*} & \text{if } n^{o^*}_p \leq n^{o^*}_v, \\
n^{o^*}_v & \text{otherwise,}
\end{cases}
\end{equation}
where $n_p^{o^*} < n_v^{o^*}$ indicates that excess instances should be removed, and $n^{o^*}_p > n^{o^*}_v$ suggests that additional instances are required.
For example, in \cref{fig:method}, if $O^*$ represents people with $n^{o^*}_p=2$ and $n^{o^*}_v=1$, we set $N^*=1$ to preserve one person.  
Note that multiple instances of $O^*$ may exist if there are several plausible entities to retain.  
For \textit{global} scene elements (e.g., background), which are not inherently countable, 
we treat $N^*$ as a presence indicator, setting $N^*=1$ if the element is preserved.  
This unified notation allows us to consistently handle both entity- and scene-level preservation within the same refinement planning framework. (see Appendix \cref{sec:appendix-global-refinement} for global refinement performance).

\paragraph{Identifying regions to preserve.}
Based on the entity selection $O^*$, we localize the regions corresponding to correctly generated content within the video frames, as shown in \cref{fig:method} (top right). 
For countable entities, given the set of preserved instances $O^*$ and their quantities $N^*$, we first construct a pointing prompt using the template: “Point the biggest \{$N^*$\} \{$O^*$\}” (e.g., “Point the biggest 1 bear”). This prompt is used to obtain 2D coordinates indicating the spatial locations of $O^*$ in each sampled frame. Using these coordinates as initialization, we apply a segmentation model to extract entity-specific regions, resulting in binary segmentation masks $\mathbf{M} \in \mathbb{R}^{H \times W \times T}$ that preserve the correctly generated entities.
In practice, for global elements such as background, we simply preserve the entire frame region or assign a broad background mask if the property is rendered correctly. By combining these with the region-level masks, we obtain a dense, frame-aligned segmentation map $\mathbf{M}$ that preserves both entity- and scene-level regions.

\paragraph{Prompt regeneration for regions requiring refinement.} 
We additionally generate a local prompt for refinement to enable distinct control over different regions during generation. To this end, we prompt an LLM to produce a refinement-oriented prompt, $p^r$, based on $Q$ but excluding any questions related to $O^*$. 
As illustrated in \cref{fig:method}, this regenerated local prompt will be used to guide the denoising process for specific areas to be refined during video generation in a later stage.

\subsection{\colorbox{step3}{Localized Refinement}}\label{sec:subsec:localized_refinement}

At this stage, we refine the video to improve alignment while preserving coherence with the original content.
While video editing~\citep{jiang2025vace,yang2025videograin} preserves masked regions and enforces visual consistency, it is limited in its ability to freely introduce or correct entities misaligned with the original prompts.
Similarly, inpainting~\citep{lugmayr2022repaint,bian2025videopainter} fills missing regions with locally consistent textures but lacks mechanisms for semantically guided object introduction or correction from textual input. (See \Cref{tab:ablation_components})
Instead of these approaches, we selectively re-initialize noise only in misaligned regions and apply distinct text prompts to preserved and refined areas, enabling targeted corrections while maintaining overall video consistency.

\paragraph{Localized noise re-initialization.}
We adopt a mask-based strategy in which only regions marked for refinement are re-initialized with newly sampled noise $\epsilon'_0 \sim \mathcal{N}(0, \mathbf{I})$, while preserved regions retain their original noise $\epsilon_0$.
This selective resampling maintains consistency in faithful areas while allowing controlled updates in misaligned ones.
To transform the pixel-level mask $\mathbf{M}$ into the latent space, we apply block averaging (pooling), yielding a hybrid noise map:
\begin{equation}
\begin{aligned}
\epsilon^*_0 =& \left(\epsilon_0 \otimes \text{pool}(\mathbf{M}, d)\right) \\
+ & \left(\epsilon'_0 \otimes (1 - \text{pool}(\mathbf{M}, d))\right)
\end{aligned}
\end{equation}
where $\text{pool}(\cdot, d)$ downsamples the mask and $\otimes$ denotes element-wise multiplication. 
This noise map $\epsilon_0^*$ is then used with localized prompts to guide the frozen diffusion model.

\paragraph{Localized text guidance.}
Afterward, we apply distinct text prompts to different spatial regions of the video based on their noise re-initialization status, using the binary segmentation mask $\mathbf{M}$ to separate preserved ($M_{\text{pres}}$) and re-initialized ($M_{\text{refine}} = 1 - M_{\text{pres}}$) areas. 
For the re-initialized regions, we guide generation in the latent space using regenerated prompts $p^r$ (See~\cref{sec:subsec:planning}) tailored to those areas. 
In parallel, motivated by recent findings on noise bias~\citep{noise2, noise3, noise4}, we reuse the original prompt $p$ to preserve features related to $O^*$ in the retained regions. 
This regionalized decomposition of the original prompt allows for the addition or modification of objects in re-initialized areas, while maintaining the integrity of correctly generated content in preserved regions.

\paragraph{Harmony with original elements.} 
To further ensure global coherence between preserved and refined regions, 
we regenerate all pixels through two separate diffusion paths and fuse them via joint optimization. 
Specifically, at each denoising step $t$, we run the diffusion model $f(\cdot)$ twice with different prompts and noises: 
$\hat{V}_{\text{pres}} = f(V_t, p, \epsilon_0)$ for the preserved regions, and 
$\hat{V}_{\text{refine}} = f(V_t, p^r, \epsilon_0')$ for the refined regions. 
The final fused output $\tilde{V}$ is obtained by solving:
\begin{equation}
\begin{aligned}
V_1 = \arg\min_{\tilde{V}} \;&
\left\| M_{\text{pres}} \otimes (\tilde{V} - \hat{V}_{\text{pres}}) \right\|^2 \\
&+ \left\| M_{\text{refine}} \otimes (\tilde{V} - \hat{V}_{\text{refine}}) \right\|^2
\end{aligned}
\end{equation}

This joint optimization allows $\tilde{V}$ to seamlessly blend preserved and refined regions, 
reducing mismatches at region boundaries and producing perceptually smooth, globally coherent videos.

\input{table/evalcrafter}

\paragraph{Video ranking.} 
Similar to generating multiple candidate prompts in~\citep{opt2i}, we produce $K$ refined videos using different random seeds and select the best one based on our video scores, as obtained in~\cref{sec:subsec:video_evaluation}, thus avoiding additional computations or resource burdens. If multiple videos receive a tied video score, 
we select the video with the highest BLIP-BLEU score~\citep{liu2023evalcrafter} among them.

%% file: table/evalcrafter.tex
\begin{table*}[t]
\centering
\footnotesize
\caption{\textbf{Evaluation results on EvalCrafter with 
other baselines.
} 
Note that we focus on these four splits, whereas the official website reports the average across all splits.
We highlight the quality and consistency performance in \textcolor{red}{red} if it deteriorates by more than 1\% from the original performance.
}
\label{tab:results_evalcrafter}
\renewcommand{\arraystretch}{0.9}
\resizebox{.99\textwidth}{!}{
\begin{tabular}{l ccccc c c c }

\toprule
\multirow{2}{*}{Method} & \multicolumn{5}{c}{Text-Video Alignment} & 
\multirow{2}{*}{\begin{tabular}{c}
    Visual \\
    Quality   
\end{tabular}} & 
\multirow{2}{*}{\begin{tabular}{c}
     Motion\\
     Quality
\end{tabular}} 
& \multirow{2}{*}{\begin{tabular}{c}
    Temporal\\
    Consistency
\end{tabular}} 

\\

\cmidrule(lr){2-6}
 & Count & Color & Action & Others & Avg.  \\ \midrule
\rowcolor[HTML]{EFEFEF} 
VideoCrafter2 & 47.52 & 46.28 & 44.07 & 46.02 & 45.97 & \textcolor{gg}{61.8} & \textcolor{gg}{62.6} & \textcolor{gg}{62.9}  \\ 
+ LLM paraphrasing & 45.87 & 47.81 & 44.41 & 45.16 & 45.81 & \textcolor{bblue}{62.4} & \textcolor{bblue}{62.7} & \textcolor{bblue}{62.7}  \\ 
+ SLD & 44.47 & 46.45 & 39.89 & 44.06 & 43.72  & \textcolor{red}{52.5} &  \textcolor{bblue}{62.2} & \textcolor{red}{44.4}  \\
+ OPT2I & 47.69 & 47.67 & 45.04 & 44.65 & 46.26 & \textcolor{bblue}{62.1} & \textcolor{bblue}{62.6} & \textcolor{bblue}{62.8}  \\ 
\rowcolor{blue!8} 
+ \ours{} (Ours) & \textbf{49.84} & \textbf{51.57}& \textbf{45.78} & \textbf{48.12} & \textbf{48.83} & \textcolor{bblue}{62.1} & \textcolor{bblue}{62.4} &  \textcolor{bblue}{62.0} \\ 

\midrule
\rowcolor[HTML]{EFEFEF} 
T2V-turbo & 46.14 & 43.06 & 41.42 & 43.16 & 43.94 & \textcolor{gg}{63.3} & \textcolor{gg}{57.8} & \textcolor{gg}{61.6} \\
+ LLM paraphrasing & 49.49 & 43.16 & 41.32 & 44.75 &44.68 & \textcolor{bblue}{62.9} & \textcolor{red}{52.9} & \textcolor{bblue}{61.9} \\ 
+ SLD & 47.39 & 43.99 & 42.13 & 43.28 & 44.20 & \textcolor{red}{56.6} & \textcolor{bblue}{58.2} & \textcolor{red}{49.2} \\ 
+ OPT2I & 47.44 & 45.00 & 44.64 & \textbf{45.54} & 45.66 & \textcolor{bblue}{63.3} & \textcolor{red}{56.4} & \textcolor{red}{48.9} \\ 
\rowcolor{blue!8} 
+  \ours{} (Ours) & \textbf{51.27} & \textbf{46.66} & \textbf{45.81} & 45.45 &\textbf{47.30} & \textcolor{bblue}{63.2} & \textcolor{bblue}{57.9}  & \textcolor{bblue}{61.8} \\ 

\midrule
\rowcolor[HTML]{EFEFEF} 
CogVideoX-5B & 47.88  &  49.63 & 37.76 & 44.78 & 45.01 & \textcolor{gg}{65.8} & \textcolor{gg}{61.0} & \textcolor{gg}{61.8} \\ 
+ LLM paraphrasing & 45.58 & 46.56 & 37.17 & 43.18 & 43.12 & \textcolor{red}{58.4} & \textcolor{bblue}{61.1} & \textcolor{bblue}{61.7} \\ 
+ SLD & 47.73 & 46.27 & 39.55 & 43.75 & 44.33 & \textcolor{red}{49.6} & \textcolor{red}{51.2} & \textcolor{red}{21.0} \\ 
+ OPT2I & 48.62 & 48.89 & \textbf{41.39} & 43.62 & 45.63 & \textcolor{red}{59.7} & \textcolor{bblue}{60.9}&  \textcolor{bblue}{61.9} \\ 
\rowcolor{blue!8} 
+ \ours{} (Ours) & \textbf{49.63} & \textbf{49.94} & 40.69 & \textbf{45.36} & \textbf{46.41} & \textcolor{bblue}{64.8} & \textcolor{bblue}{61.1} & \textcolor{bblue}{61.9} \\ 

\midrule
\rowcolor[HTML]{EFEFEF} 
Wan 2.1-1.3B & 45.06 & 48.18 & 41.06 & 45.00 & 44.83 & 63.2 & 61.0 & 62.1 \\ 
+ LLM paraphrasing & 44.38 & 47.19 & 42.35 & 44.03 & 44.49 & 63.5 & 61.2 & 62.2   \\ 
+ SLD & 48.24 & 49.77 & 43.64 & 46.78 & 47.11 & \textcolor{red}{49.1} & \textcolor{red}{58.3} & \textcolor{red}{32.5} \\ 
+ OPT2I & 49.10 & 51.86 & 45.88 & 47.92 & 48.69 & 64.3 & 61.0 & 62.0 \\ 
\rowcolor{blue!8} 
+ \ours{} (Ours) & \textbf{50.03} & \textbf{51.97} & \textbf{46.01} & \textbf{48.30} & \textbf{49.01} & 65.1 & 61.6 & 62.0 \\

\bottomrule
\end{tabular}
}
\end{table*}

%% file: sec/05_experiments.tex
\section{Experiments}
\label{sec:experiments}

\begin{figure*}[t]
    \centering
    {\includegraphics[width=0.99\linewidth]{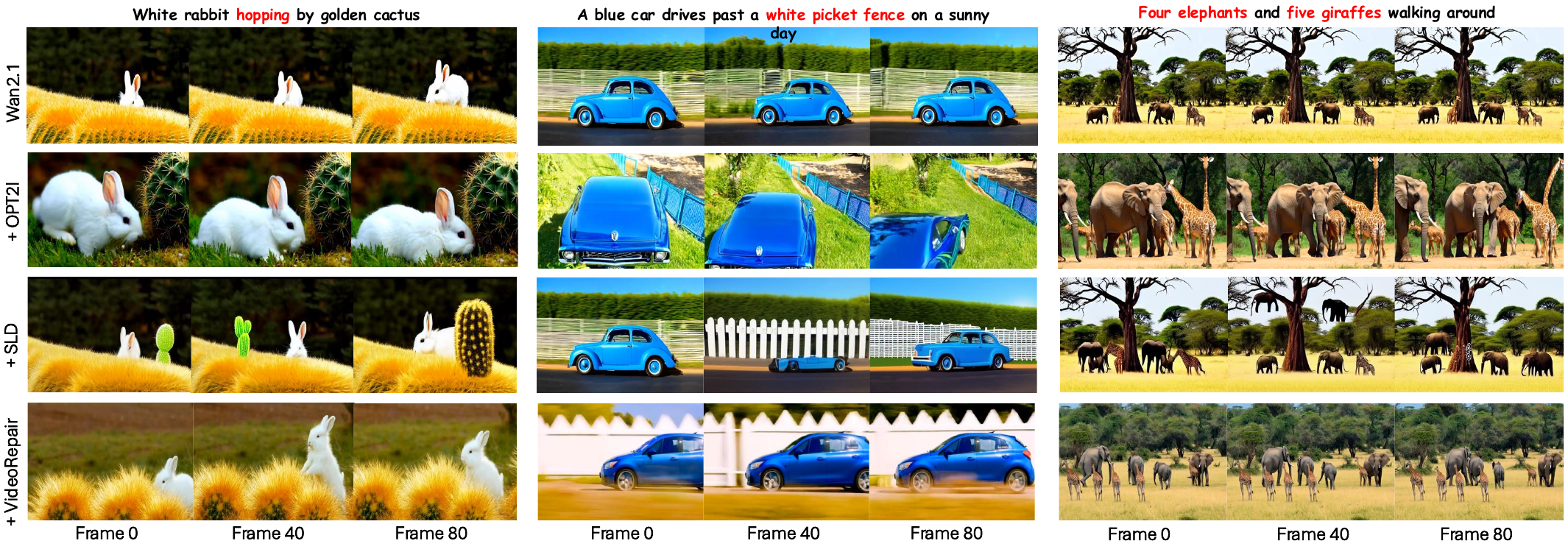}}
    \caption{ 
    \textbf{Comparison of refinement framework with \ours{} (backbone: Wan2.1-1.3B).}
    \ours{} successfully refines misaligned motion, adds missing components, and corrects numeracy.  
    }
    \label{fig:main_qualitative}    
\end{figure*}

\subsection{Experiment Setups}
\label{sec:exp_setups}

\paragraph{Benchmarks and evaluation metrics.}

We evaluate our method on two T2V generation benchmarks:
EvalCrafter~\citep{liu2023evalcrafter} and T2V-CompBench~\citep{sun2024t2v}.\footnote{All reported results are based on our own experiments. We use ver.1 of T2V-CompBench.}
Details are provided in the Appendix \cref{sec:appendix-evaluation-details}.

\textbf{(1) EvalCrafter.} 
We follow the official metadata 
to split prompts by attributes into four sections: \textit{count}, \textit{color}, \textit{action}, and \textit{others}. 
The \textit{others} category includes scenery-related prompts, such as Camera movement (e.g., \textit{"Zoom in"}), 
Landscape (e.g., \textit{"A bustling street in Paris"}), and Style (e.g., \textit{"Polaroid style"}). 
For evaluation metrics, we report four groups: text-video alignment, video quality, motion quality, and temporal consistency.

\textbf{(2) T2V-CompBench.} 
We adopt three compositional reasoning categories from this benchmark: 
spatial relationships, generative numeracy, and consistent attribute binding.
We use ImageGrid-LLaVA~\citep{liu2024improved} for consistent attribute binding evaluation 
and GroundingDINO~\citep{liu2023grounding} for the other two dimensions.

\input{table/t2v_compbench}

\paragraph{Implementation details.}
We apply \ours{} on four recent T2V models: 
T2V-turbo~\citep{li2024t2v}, VideoCrafter2~\citep{chen2024videocrafter2}, CogVideoX-5B~\citep{yang2024cogvideox}, and Wan2.1-1.3B~\citep{wan2025wan}. 
T2V-Turbo and VideoCrafter2 generate 16 frames, while CogVideoX-5B and Wan2.1 generate 81 frames.
All experiments use $K$ as 5 with a single iteration. 
For MLLM and LLM, we primarily use GPT-4o 
and for pointing and segmentation, we employ MolmoE-1B~\cite{deitke2024molmo} and Semantic-SAM (L)~\cite{li2023semantic}. 
Additional details are provided in the Appendix \cref{sec:appendix-implement-detail}.

\paragraph{Baselines.} 
We compare \ours{} against recent refinement frameworks, OPT2I~\citep{opt2i} and SLD~\citep{sld}, 
on the same three T2V models described above. 
Although these baselines were originally proposed for image refinement, 
we extend their implementations to the video setting. 
For OPT2I, we score the videos using the original DSG~\citep{dsg} and iteratively generate five prompt candidates. 
For SLD, since its refinement model is based on LMD+~\citep{lian2023llm}, 
we apply SLD frame-by-frame to the initial outputs of T2V models. 
We also include LLM paraphrasing as a baseline, where GPT-4 generates diverse paraphrases of the initial prompt. 
To ensure fairness, we unify random seeds across all experiments so that all methods refine the same initial videos. 
Further details are provided in the Appendix \cref{sec:appendix-baseline-details}.

\subsection{Quantitative Results}
\label{sec:eval_quant}

As shown in \Cref{tab:results_evalcrafter} and \Cref{tab:results_t2v_compbench}, 
\ours{} consistently outperforms other refinement baselines as well as strong compositional T2V models (e.g., Vico, VideoTetris) across both benchmarks. 
On EvalCrafter, \ours{} achieves relative alignment gains of \textbf{+6.22\%}, \textbf{+7.65\%}, \textbf{+3.11\%} and \textbf{+9.32\%} over VideoCrafter2, T2V-turbo, and CogVideoX-5B, Wan2.1 respectively. 
In addition, \ours{} effectively corrects misalignments while preserving visual fidelity (std.\ deviation of visual quality scores: $0.55$).
By contrast, SLD underperforms particularly in the \textit{action} and \textit{count} categories because its frame-level latent fusion fails to maintain consistent object counts and spatial layouts over time. 
Although OPT2I yields only modest improvements, it operates solely in the textual domain without visual guidance, which limits its ability to correct fine-grained spatiotemporal localized misalignments. Moreover, its iterative search procedure, involving multiple LLM calls, makes the refinement process computationally expensive (will be discussed in \cref{sec:ablations}).

\subsection{Qualitative Results}
\label{sec:eval_qual}
\cref{fig:main_qualitative} presents qualitative comparisons of refinement frameworks (OPT2I, SLD, and \ours{}) applied to the Wan2.1 backbone.
In the leftmost example, \ours{} preserves the \textit{golden cactus} from the initial video while refining the \textit{white rabbit’s motion} to a hopping action.
In the middle example, \ours{} maintains the \textit{blue car} and more clearly introduces the \textit{white picket fence} compared to the initial video.
In the rightmost example, \ours{} preserves the \textit{two elephants} from the initial video while correcting object numeracy during refinement.
Overall, these examples demonstrate that \ours{} effectively refines misaligned motion, introduces missing components, and corrects numerical inconsistencies.

\subsection{Additional Analysis}
\label{sec:ablations}
\begin{figure}[t]
    \centering
    \includegraphics[width=0.99\linewidth]{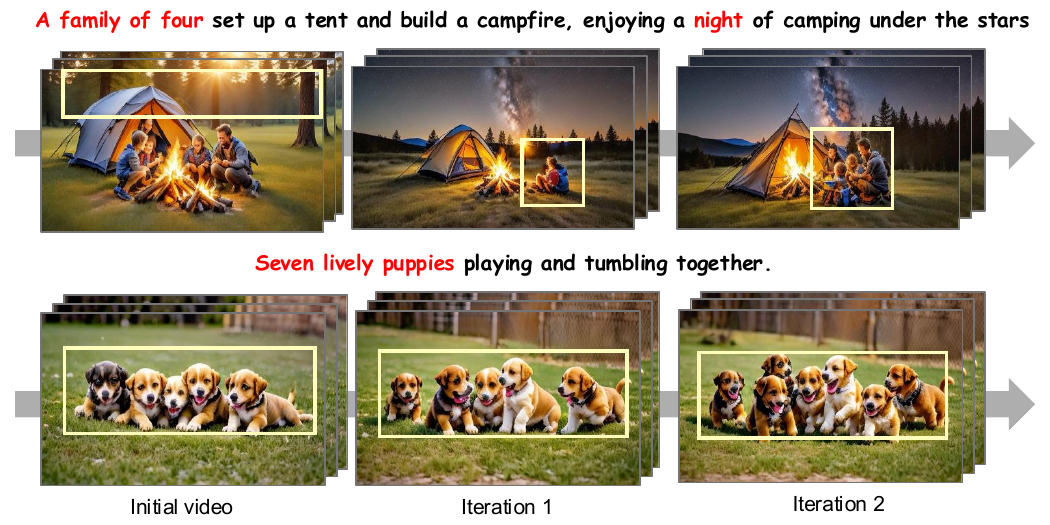}
    \caption{
\textbf{Iterative refinement. } \ours{} progressively improves text–video alignment, correcting object counts and attributes.
    }
    \label{fig:qualitative_iteration}
\end{figure}

In this section, we present additional analyses of \ours{}, including ablations of each component (\cref{tab:ablation_components,tab:error_accum,tab:ablation_step3}), error analysis (\cref{tab:failure-breakdown,tab:error-propagation}), inference latency (\cref{fig:latency}), and iterative refinement results (\cref{fig:qualitative_iteration}).
In each table, we report the average text–video alignment performance (Avg.) across the \textit{Count}, \textit{Color}, and \textit{Action} splits of EvalCrafter using the T2V-turbo backbone.
Our default setup is highlighted with a purple background.

\paragraph{Ablations of each step’s components.}
In \cref{tab:ablation_components}, we analyze the contributions of individual components, including the evaluation question type, planning strategy, and ranking metric. 
Since \ours{} operates as a sequential pipeline, we ablate each component by replacing it with an alternative while keeping all other steps unchanged.
For evaluation questions, we compare the original DSG questions with our proposed questions. For the object selection, we evaluate random planning against our strategy guided by previous evaluation results. For the ranking stage, we compare different metrics for final video selection, including CLIP, BLIP-BLEU, and our ranking method.
Comparing with just applying ranking on T2V-turbo generations (44.64), \ours{} achieves a notable improvement (47.91).

\paragraph{Ablations of localized refinement models.}
In \cref{tab:ablation_step3}, we evaluate the effectiveness of our localized refinement by comparing it with state-of-the-art video-to-video (V2V) editing models, including VACE~\citep{jiang2025vace} and VideoGrain~\cite{yang2025videograin}, using our planning outputs ($M$ and $p^r$) as editing guidance.
Specifically, using $M$ alone corresponds to masked V2V editing, while using both $M$ and $p^r$ incorporates $p^r$ as an additional textual prompt during masked V2V editing.
Although masked V2V editing can partially resemble our localized refinement, its editing capability is limited to modifying attributes or shapes of correctly generated objects.
In contrast, our localized refinement generalizes to diverse failure cases by identifying and correcting various types of misaligned objects and errors. More examples are provided in \cref{fig:qual_scale_category_examples}.

\paragraph{Ablations of MLLM components.}
We analyze the impact of replacing \ours{}'s components across different stages (video evaluation, planning, and ranking) with human annotations and various MLLMs (GPT-4o, Qwen2.5VL-7B, and Gemini-2.5-Flash). 
As shown in \cref{tab:error_accum}, \ours{} remains robust under diverse model substitutions, with only minimal performance variation, highlighting the flexibility and modularity of the framework.
Using human evaluators for video evaluation and planning provides a strong upper bound, while replacing them with GPT-4o results in only a modest degradation, indicating limited error accumulation when using strong proprietary models. 
Furthermore, mixing different models across stages can be beneficial: combining complementary strengths of models (e.g., Qwen2.5VL-GPT4o-Gemini) maintains or even slightly improves performance. 
This suggests that \ours{} not only tolerates heterogeneous components but can also leverage them synergistically, enabling flexible deployment under varying resource constraints.

\input{table/ablation_components}

\input{table/error_accum_latency}

\input{table/step3_ablation}

\begin{table}[t]
\centering
\caption{\textbf{Categorized breakdown of failures.} We analyze failure modes across different stages of the pipeline and report their occurrence percentages.}
\resizebox{\linewidth}{!}{%
\begin{tabular}{cc|c}
\toprule
\textbf{Category} & \textbf{Error Type} & \textbf{Percentage (\%)} \\
\midrule
 \cellcolor[HTML]{E7F2DC}Misalignment Detection & QA hallucination & 35.3 \\
\midrule
\cellcolor[HTML]{FBF5DC}Refinement Planning   & Planning error   & 11.8 \\
\midrule
\cellcolor[HTML]{FBE4E7} & Mask drift       & 58.8 \\
 \cellcolor[HTML]{FBE4E7}Localized Refinement  & Boundary artifacts & 23.5 \\
\cellcolor[HTML]{FBE4E7} & Identity inconsistency & 47.1 \\
\bottomrule
\end{tabular}
}
\label{tab:failure-breakdown}
\end{table}

\paragraph{Inference latency.}
\cref{fig:latency} presents the relationship between text–video alignment performance and inference latency across different refinement frameworks.
OPT2I and SLD incur 3–5× higher inference latency compared to \ours{}, without achieving superior alignment performance. OPT2I requires an iterative prompt search procedure, involving multiple rounds of video generation and DSG-based evaluation. Similarly, SLD depends on an external generator to synthesize missing objects and performs frame-wise copy-and-paste operations, resulting in additional computational overhead.
In contrast, \ours{} achieves the best trade-off between text–video alignment and efficiency, achieving strong performance with relatively modest inference cost.
Step-wise inference latency and detailed cost comparisons are provided in Appendix \cref{tab:inference-latency,tab:cost}.

\paragraph{Iterative refinement.}
We further explore iterative refinement to progressively enhance text-video alignment, as a single refinement step of \ours{} may not fully resolve all inconsistencies with the prompt. 
As illustrated in~\cref{fig:qualitative_iteration}, the first refinement partially corrects the misalignment by generating a scene of \textit{a night of camping under the stars}, but some family members disappear. 
In the second iteration, \ours{} recovers all four family members while preserving the rest of the scene. 
Similarly, in the bottom example of~\cref{fig:qualitative_iteration}, iterative refinement successfully produces the intended output of seven puppies. 
Additional qualitative examples are provided in the Appendix~\cref{sec:appendix-iter-refinement}.

\begin{table}[t]
\centering
\small
\caption{\textbf{Error propagation analysis.} Conditional probabilities of downstream errors given upstream failures.}
\begin{tabular}{cc|c}
\toprule
\textbf{Upstream Error} & \textbf{Downstream Error} & \textbf{Prob. (\%)} \\
\midrule
\cellcolor[HTML]{E7F2DC}QA hallucination & \cellcolor[HTML]{FBF5DC}Planning error & 16.7 \\
\cellcolor[HTML]{E7F2DC}QA hallucination & \cellcolor[HTML]{FBE4E7}Mask drift & 21.3 \\
\cellcolor[HTML]{E7F2DC}QA hallucination & \cellcolor[HTML]{FBE4E7}Identity inconsistency & 9.8 \\
\midrule
\cellcolor[HTML]{FBF5DC}Planning error & \cellcolor[HTML]{FBE4E7}Mask drift & 20.5 \\
\cellcolor[HTML]{FBF5DC}Planning error & \cellcolor[HTML]{FBE4E7}Identity inconsistency & 10.3 \\
\midrule
\cellcolor[HTML]{FBE4E7}Mask drift & \cellcolor[HTML]{FBE4E7}Identity inconsistency & \textbf{43.5} \\
\cellcolor[HTML]{FBE4E7}No mask drift & \cellcolor[HTML]{FBE4E7}Identity inconsistency & 28.6 \\
\bottomrule
\end{tabular}
\label{tab:error-propagation}
\end{table}

\paragraph{Categorized failure analysis.}
To better understand failure modes, we randomly sample 50 failure cases and conducted a manual categorization analysis. We identify five representative error types: \textit{QA hallucination} (incorrect misalignment detection), \textit{planning error} (incorrect object selection), mask drift (frame-wise segmentation inaccuracies), \textit{boundary artifacts} (unnatural blending), and \textit{identity/motion inconsistencies} in preserved regions. Since multiple errors co-occur in a single case, percentages do not sum to 100\%.
As shown in \cref{tab:failure-breakdown}, mask drift (58.8\%) and identity/motion inconsistencies (47.1\%) are the most frequent failure modes. Mask drift primarily arises from frame-level segmentation variability under occlusion. Identity/motion inconsistencies stem from the joint optimization formulation: although preserved regions retain original noise initialization and conditioning, they are not strictly frozen and may slightly adjust during least-squares fusion, leading to subtle perceptual shifts.

\paragraph{Error propagation analysis.}
To quantitatively assess potential error propagation, we analyze conditional probabilities $P(\text{Downstream Error}|\text{Upstream Error})$ based on the manually categorized failure cases described above.
As shown in \cref{tab:error-propagation}, \textit{error propagation is neither deterministic nor systematically cascading across stages.} QA hallucinations rarely propagate downstream, as we conservatively preserve only confidently verified objects and filter out spurious detections. Planning errors are similarly weakly coupled with mask-level failures: even when an incorrect object is selected, refinement proceeds consistently with that selection rather than amplifying segmentation instability.
In contrast, mask drift exhibits the strongest downstream effect, emerging as the primary driver of identity or motion inconsistencies during refinement.

%% file: table/t2v_compbench.tex
\begin{table}[t]
\centering
\footnotesize
\caption{\textbf{Evaluation results on T2V-CompBench.}}
\label{tab:results_t2v_compbench}
\resizebox{\linewidth}{!}{%
\begin{tabular}{l cccc}
\toprule
Method & Consist-Attr & \multicolumn{1}{c}{Spatial} & Numeracy & Avg. \\ 
\midrule
ModelScope & 0.5148 & 0.4118 & 0.1986 & 0.3750 \\
ZeroScope & 0.4011 & 0.4287 & 0.2408 & 0.3568 \\
Latte & 0.4713 & 0.4340 & 0.2320 & 0.3791 \\
Show-1 & 0.5670 & 0.4544 & 0.3086 & 0.4433 \\
Open-Sora-Plan & 0.4246 & 0.4520 & 0.2331 & 0.3699 \\ 
Vico & 0.6470 &0.5425  & 0.2762 & 0.4886\\
VideoTetris & 0.6211 & 0.4832 & 0.3467 & 0.4836 \\ \midrule
VideoCrafter2 & 0.6812 & 0.5214  & 0.2906 & 0.4977 \\
 \rowcolor{blue!8} 
  + \ours{} & \textbf{0.7275}  & \textbf{0.5690}  & \textbf{0.3278} & \textbf{0.5383} \\ 
  \midrule

T2V-turbo & 0.7025 &  0.5492 & 0.2496 & 0.5004 \\
\rowcolor{blue!8} 
  + \ours{} & \textbf{0.7675}  & \textbf{0.5807}  & \textbf{0.2709} & \textbf{0.5439}  \\
  \midrule

CogVideoX-5B & 0.6220  & 0.4988  & 0.2228 & 0.4479 \\
 \rowcolor{blue!8} 
+ \ours{} & \textbf{0.6725}  & \textbf{0.5811}  & \textbf{0.3034} & \textbf{0.5190}  \\ 
\midrule

Wan2.1-1.3B & 0.6870  & 0.5690 & 0.3516 & 0.5358 \\
 \rowcolor{blue!8} 
+ \ours{} & \textbf{0.7262}  & \textbf{0.5841} & \textbf{0.3837} & \textbf{0.5646}  \\

\bottomrule
\end{tabular}
}
\end{table}

%% file: table/ablation_components.tex
\begin{table}[t]
\centering
\caption{
\textbf{Ablations of \ours{} components.} We replace each stage while keeping others fixed. 
}
\renewcommand{\arraystretch}{0.85}
\label{tab:ablation_components}
\resizebox{\linewidth}{!}{%
\begin{tabular}{c c c | c}
\toprule
\begin{tabular}[c]{@{}c@{}}\cellcolor[HTML]{E7F2DC}Question type\\ \cellcolor[HTML]{E7F2DC}(\cref{sec:subsec:video_evaluation})\end{tabular} &
\begin{tabular}[c]{@{}c@{}}\cellcolor[HTML]{FBF5DC}Planning\\ 
\cellcolor[HTML]{FBF5DC}(\cref{sec:subsec:planning})\end{tabular} &
\begin{tabular}[c]{@{}c@{}}\cellcolor[HTML]{FBE4E7}Ranking metric\\ \cellcolor[HTML]{FBE4E7}(\cref{sec:subsec:localized_refinement})\end{tabular} &
Avg. \\
\midrule
- & - & - & 43.54 \\ 
- & - & Ours & 44.68 \\ 
\midrule
DSG & Random & Ours & 45.18 \\
Ours & Random & Ours & 46.92 \\ 
\midrule
Ours & Ours & CLIP & 45.85 \\
Ours & Ours & BLIP-BLEU & 47.77 \\
\rowcolor{blue!8}
Ours & Ours & Ours & \textbf{47.91} \\
\bottomrule
\end{tabular}
}
\vspace{.1in}
\end{table}

%% file: table/error_accum_latency.tex
\begin{table}[t]
\centering
\small
\caption{\textbf{Robustness under MLLM substitution.} We replace components in each stage with different MLLMs and report the average T2V alignment.}
\label{tab:error_accum}
\resizebox{\linewidth}{!}{%
\begin{tabular}{ccc|c}
\toprule
\rowcolor[HTML]{E7F2DC} Misalign Det. & \cellcolor[HTML]{FBF5DC} Planning & \cellcolor[HTML]{FBE4E7} Ranking & \cellcolor[HTML]{FFFFFF} Avg. \\
\rowcolor[HTML]{E7F2DC} (\cref{sec:subsec:video_evaluation}) & \cellcolor[HTML]{FBF5DC} (\cref{sec:subsec:planning}) & \cellcolor[HTML]{FBE4E7} (\cref{sec:subsec:localized_refinement}) & \cellcolor[HTML]{FFFFFF}  \\
\midrule
Human & Human & GPT-4o & 49.05 \\
Human & GPT-4o & GPT-4o & 48.46 \\
Human & GPT-4o & Gemini-2.5-Flash & \textbf{49.08} \\
\midrule
\rowcolor{blue!8} GPT-4o & GPT-4o & GPT-4o & 47.91 \\
Qwen2.5VL-7B & GPT-4o & GPT-4o & 48.61 \\
Qwen2.5VL-7B & GPT-4o & Gemini-2.5-Flash & 48.79 \\
\bottomrule
\end{tabular}
}
\end{table}

%% file: table/step3_ablation.tex
\begin{figure}[t]
\centering

\begin{minipage}[c]{0.46\linewidth}
\centering
\captionof{table}{\textbf{Ablations of \\ localized refinement.}}
\label{tab:ablation_step3}
\renewcommand{\arraystretch}{1.1}
\resizebox{0.99\linewidth}{!}{%
\begin{tabular}{cc|c|c}
\toprule
\multicolumn{2}{c|}{\textit{Guidance}} & \multirow{2}{*}{Model} & \multirow{2}{*}{Avg.} \\
$M$ & $p^r$ &  &  \\ \hline
\checkmark &  & VideoGrain & 40.52 \\
\checkmark &  & VACE & 46.77 \\
\checkmark & \checkmark & VACE & 44.88 \\
\rowcolor{blue!8}
\checkmark & \checkmark & Ours & \textbf{47.91} \\ 
\bottomrule
\end{tabular}
}
\end{minipage}
\hfill
\begin{minipage}[c]{0.50\linewidth}
\centering
\includegraphics[width=1.05\linewidth]{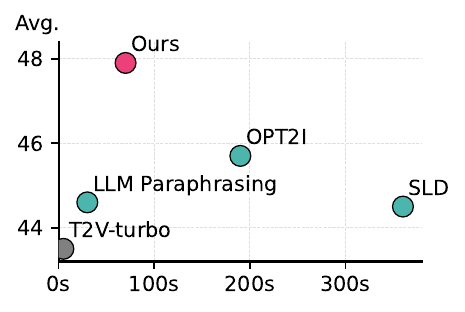}
\vspace{-0.25in}
\captionof{figure}{\textbf{Inference latency of refinement methods.}}
\label{fig:latency}
\end{minipage}

\vspace{-0.1in}
\end{figure}

%% file: sec/03_related.tex
\section{Related Works}
\paragraph{Text-to-video generation with diffusion models.}
Text-to-video (T2V) diffusion models~\citep{esser2023structure,wu2023tune,blattmann2023stable,luo2023videofusion,yang2024cogvideox,wan2025wan} aim to produce videos describing given text prompts.
VideoCrafter2~\citep{chen2024videocrafter2} synthesizes low-quality videos with high-quality images through a joint training design of spatial and temporal modules, obtaining high-quality videos. 
T2V-turbo~\citep{li2024t2v} presents a distilled video consistency model~\citep{wang2023videolcmvideolatentconsistency,song2023consistency} for improved and rapid video generation.
However, even the recent T2V diffusion models suffer from misalignment problems.
In the following,
we discuss the research direction of refining the image/video diffusion models, including \ours{}.

\paragraph{Training-free refinement for diffusion models.}
Recent works propose training-free refinement frameworks that automatically improve diffusion models' text alignment~\citep{opt2i,sld,chen2025t2i,ji2025prompt,xiang2025promptsculptor}.
In particular, OPT2I~\citep{opt2i} presents iterative prompt optimization,
where an LLM provides various variations of text prompts, T2I diffusion models generate images from the prompts, and the images are ranked with a T2I alignment score such as DSG~\citep{dsg} to provide the final image.
Since no explicit feedback is given to the backbone generation model,
it usually takes long iterations (\eg{}, 30 LLM calls) to find a prompt that provides improved alignment, making the framework expensive to use in practice.
SLD~\citep{sld} provides more explicit guidance by generating a bounding-box level plan with an LLM, followed by operations such as object addition, deletion, and repositioning.
However, SLD depends on an external layout-guided object generator (\eg{}, GLIGEN~\citep{li2023gligen}) to insert objects, and the added content often fails to harmonize with the original image.
In contrast, \ours{} is the first training-free refinement framework that delivers fine-grained localized feedback and is compatible with any T2V diffusion model, without relying on additional generators.

%% file: sec/06_conclusion.tex
\section{Conclusion}
\label{sec:conclusion}
We propose \ours{}, a training-free, model-agnostic video refinement framework that improves text-to-video alignment, including three stages: misalignment detection, refinement planning with key-object preservation, and localized refinement. \ours{} consistently outperforms recent baselines on two benchmarks, supported by extensive ablations and qualitative results.

\section{Limitation}
We note that the limitations of \ours{} primarily stem from the capability of the underlying evaluator (MLLM), rather than the refinement framework itself. First, our method is less effective for ambiguous or subjective prompts. Since \ours{} relies on explicit, semantically grounded feedback signals (e.g., discrepancies in object count, color, or action), cases without well-defined and objectively verifiable criteria make it inherently difficult to generate reliable corrective feedback.
Second, \ours{} is limited in handling prompts that require precise text rendering or fine-grained facial attributes (e.g., specific celebrity identity). This is because current MLLM-based evaluators are not sufficiently reliable for structured assessment of exact text fidelity (such as spelling accuracy) or subtle identity consistency in faces.

\section*{Acknowledgments}
We thank the anonymous reviewers for their valuable feedback and constructive suggestions.
This work was supported by DARPA ECOLE Program No. HR00112390060, NSF-AI Engage Institute DRL-2112635, DARPA Machine Commonsense (MCS) Grant N66001-19-2-4031, ARO Award W911NF2110220, ONR Grant N00014-23-1-2356, Accelerate Foundation Models Research program, and a Bloomberg Data Science PhD Fellowship. The views contained in this article are those of the authors and not of the funding agency.

%% file: sec/Appedix.tex
\clearpage
\appendix
\crefalias{section}{appendix}

\section*{Appendix}

\addcontentsline{toc}{section}{Appendix Table of Contents}
\startcontents[appendix]
\printcontents[appendix]{l}{1}{\setcounter{tocdepth}{2}}

\section{Implementation Details}\label{sec:appendix-implement-detail}

\subsection{Evaluation Question Generation}\label{sec:appendix-qa-detail}

Given the initial prompt, we construct a semantic tuple $\mathcal{T}$, similar to DSG~\citep{dsg} that contains  \textit{entities (subjects)}, \textit{attributes}, and \textit{relationships}. Here,
attributes are expressed in 2-tuples (subjects, its attribute,~(\eg{}, $\texttt{\{bed, blue\}}$), and relationships are in 3-tuples (subject entity, object entity, and their relationship, (\eg, $\texttt{\{people, pizza, make\}}$).
Based on $\mathcal{T}$, which covers all scene-relevant information, we generate questions $Q$ using \textit{GPT-4-0125}~\citep{openai2024gpt4}.
Note that although DSG does not account for object counts by design, we can incorporate assessments for whether the generated videos contain the correct number of target objects, thereby guiding automatic refinement with greater accuracy.
For example, given a prompt `there is a bear', DSG only generates an evaluation question ``is there a bear?'', which only checks the bear's existence, but does not penalize when more than one bear is generated.

\subsection{Video Object Evaluation}\label{sec:appendix-obj-eval}
\begin{figure*}[h]
    \centering
    \includegraphics[width=1.0\linewidth]{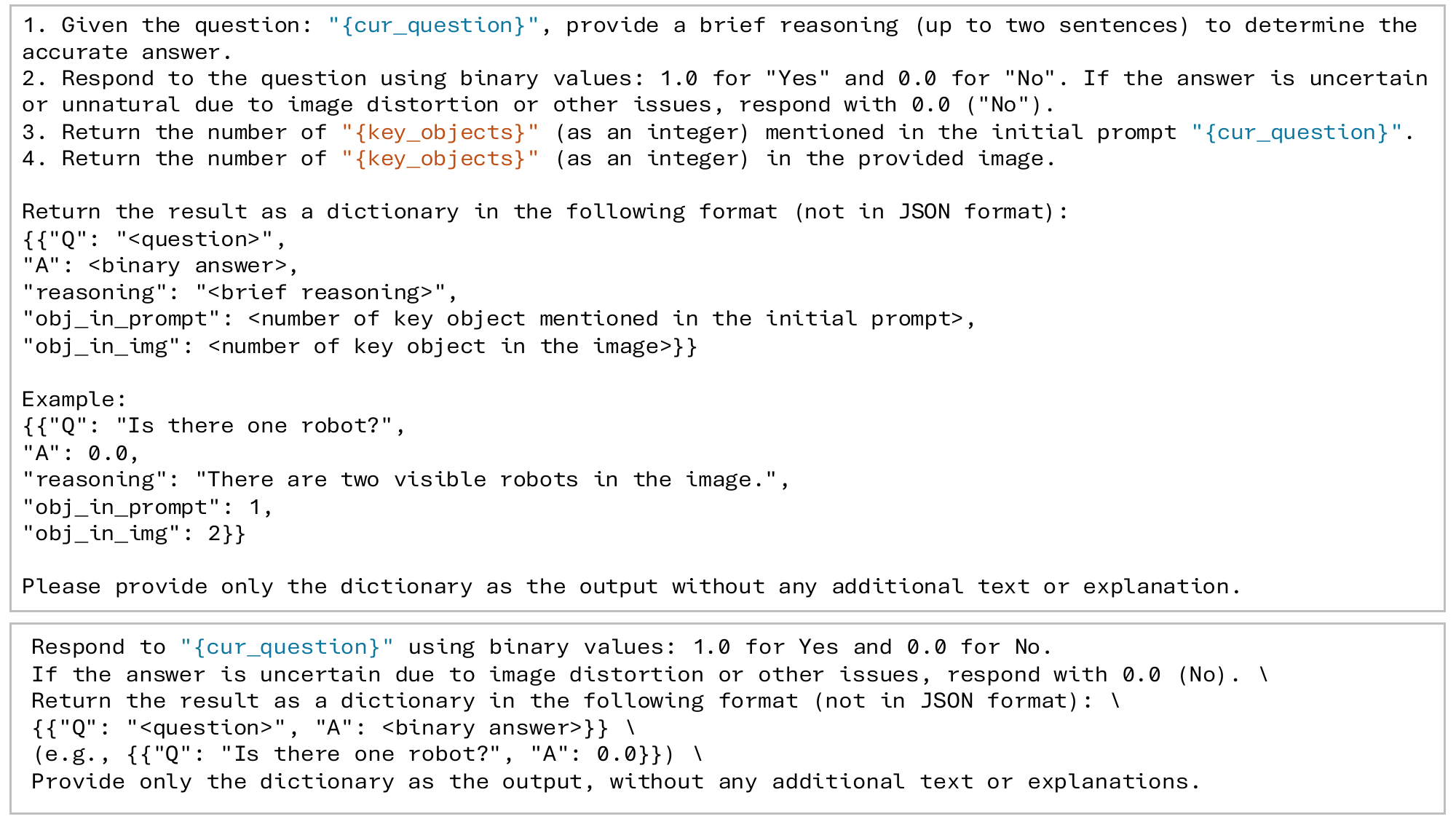}
    \caption{\textbf{Prompts to perform visual question answering in video evaluation steps.} \textbf{Top:} The prompt for $Q^o_c$ (count-related question),
    \textbf{Bottom:} prompt for $Q^o_\text{others}$ (attribute-related question). 
    \texttt{cur\_question} means each video evaluation question and \texttt{key\_objects} means entity word in each question.  
    }
    \label{fig:prompt1_vqa}
\end{figure*}

To evaluate the generated videos, we utilize GPT-4o to answer both count-related ($Q^o_c$) and attribute-related ($Q^o_a$) questions, as illustrated in \cref{fig:prompt1_vqa}. 
For $Q^o_c$ prompts, we guide GPT-4o through four steps: reasoning, answering, counting the predicted number of objects ($n^o_p$), and verifying the true count ($n^o_v$). These steps yield an answer triplet $A_c^o = \{b_c^o, n^o_p, n^o_v\}$. To ensure valid responses, we account for dependencies among questions, following the methodology of DSG~\citep{dsg}.
Each question is sequentially presented to GPT-4o, and the video score is computed as the proportion of correctly answered binary questions across all VQA tasks.
If the video score reaches 1.0 (indicating a perfect score), the \ours{} process is terminated.

\subsection{Key Object Extraction}\label{sec:appendix-key-obj}
\begin{figure*}[h]
    \centering
    \includegraphics[width=1.0\linewidth]{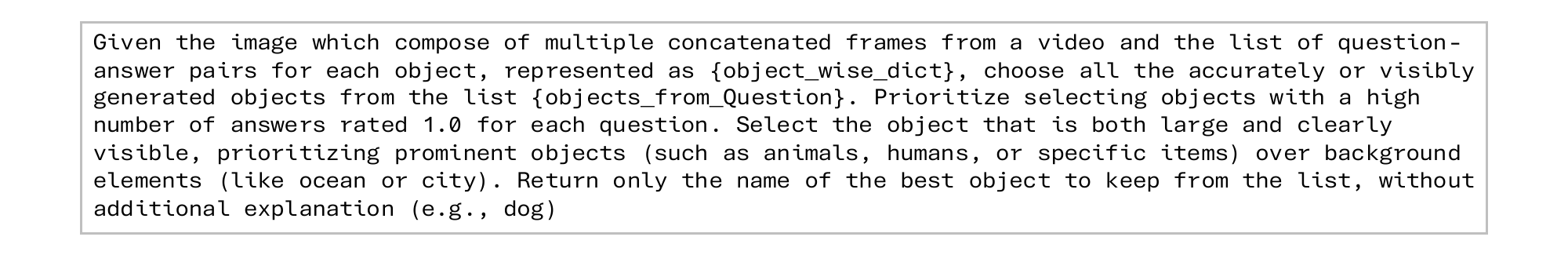}
    \caption{\textbf{Prompt to choose which object(s) to preserve.} We ask GPT4o to select objects to preserve in the scene. }
    \label{fig:prompt2_keyobj}
\end{figure*}

To extract the key concept $O^*$ from the initial videos $V_0$, we sampled frames of $V_0$ and the list of question-answer pairs for each object to GPT4o as shown in \cref{fig:prompt2_keyobj}. 
Here, we prioritize selecting objects with a higher number of 1.0 video scores. 
Moreover, we force GPT4o to select `object' instead of `background' elements to improve the accuracy of region decomposition by pointing. 

\subsection{Refinement Prompt Generation}\label{sec:appendix-refinement-prompt}
\begin{figure*}[h]
    \centering
    \includegraphics[width=1.0\linewidth]{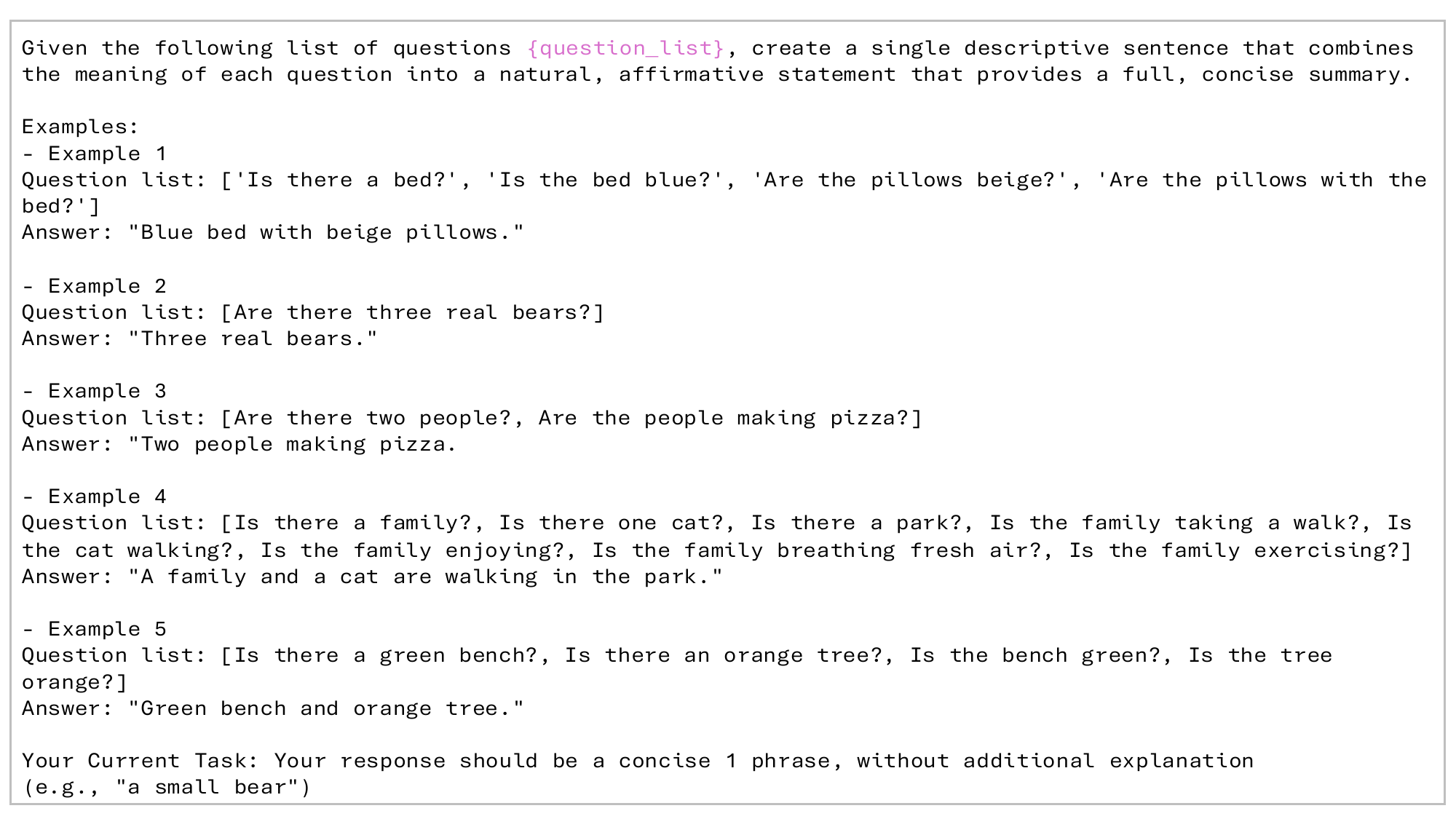}
    \caption{\textbf{Prompt to plan how to refine the other regions.} We use five in-context examples to create the refinement prompt from the question related to other objects. }
    \label{fig:prompt3_q2prompt}
\end{figure*}

To produce a refinement prompt $p^r$, we use GPT4 with instruction as shown in \cref{fig:prompt3_q2prompt}.
After getting $O^*$, we can decompose the whole question set $Q$ as $Q^{o*}$ and others depending on whether the $O^*$ keyword is included in the question.
To generate $p^r$ from specific question sets, we utilize five manually crafted in-context examples to ensure the accuracy of the generation process.
If the video score is 0.0 (indicating a complete failure from VQA) and the key object $O^*$ cannot be identified, we consider the T2V model to have failed in generating any object correctly. In such cases, we paraphrase $Q$ directly into $p^r$ using a large language model (LLM).

\section{Additional Baseline Details}\label{sec:appendix-baseline-details}

\begin{figure*}[h]
    \centering
    \includegraphics[width=.65\linewidth]{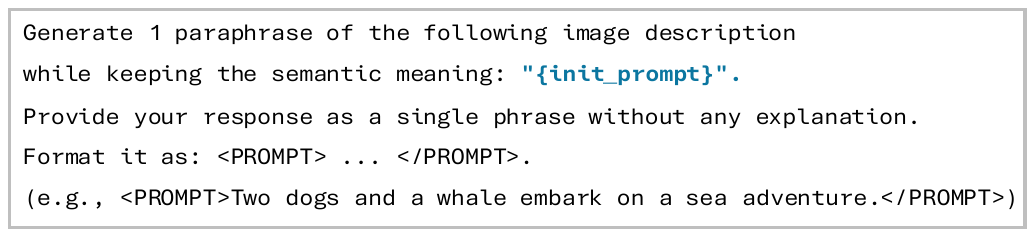}
    \caption{\textbf{Prompt for LLM paraphrasing.} Following OPT2I~\citep{opt2i}, we ask GPT4 to generate
diverse paraphrases of each prompt for LLM paraphrasing baseline experiments.}
    \label{fig:llm_para}
\end{figure*}

\paragraph{\textbf{LLM Paraphrasing.}}
Following \citep{opt2i}, we compare \ours{} with paraphrasing prompts from LLM. 
Here, we ask GPT4 to generate diverse paraphrases of each prompt, without any context about the consistency of the
images generated from it.
The prompt used to obtain paraphrases is provided in \cref{fig:llm_para}.

\paragraph{\textbf{OPT2I.}}
Since OPT2I~\citep{opt2i} aims to improve text-image consistency for T2I models, we reimplement OPT2I for T2V. 
Specifically, we replace the original T2I model part with T2V models (T2V-Turbo and VideoCrafter2) to generate outputs. Using GPT-4o, we then pose DSG questions to these outputs. For prompts, we directly adopt the ones provided in the original OPT2I paper. For LLM, we use GPT4 as \ours{}. Finally, we perform iterative refinement, running 10 iterations for T2V-Turbo and 5 iterations for VideoCrafter2, with five video candidates per iteration.

\paragraph{\textbf{SLD.}}
To adapt SLD~\citep{sld} to the T2V setup, we apply their official code to individual video frames and maintain their default setup. 
Note that SLD is a GLIGEN~\citep{li2023gligen}-based T2I model, which poses challenges for direct extension to video generation. Since SLD operates using DDIM inversion, we use the initial videos generated by T2V-Turbo and VideoCrafter2 as inputs, enabling the implementation of their noise composition method. 
Here, we use one iteration for SLD and GPT4 for LLM. 

\section{Additional Evaluation Details}\label{sec:appendix-evaluation-details}
\paragraph{EvalCrafter.}
To evaluate the effectiveness of \ours{} across different prompt dimensions, we decompose EvalCrafter~\citep{liu2023evalcrafter} using the official \texttt{metadata.json}. Specifically, we utilize the \texttt{attributes} key for each prompt and categorize the dataset into `count', `color', `action', `text', `face', and `amp (camera motion)'. 
Prompts without explicit attributes are grouped into an `others' category.
Among these dimensions, we focus on `count', `color', `action', and `others', excluding `text', `face', and `amp'. 
This decision is based on our observation that video errors related to text prompts (\eg{}, \textit{``the words `KEEP OFF THE GRASS''}), face prompts (\eg{}, \textit{``Kanye West eating spaghetti''}), and amp prompts (\eg{}, \textit{``A Vietnam map, large motion''}) cannot be reliably detected through GPT-4o question-answering, therefore hard to proceed \ours{}.

\begin{figure}[t]
    \centering
    \includegraphics[width=0.99\linewidth]{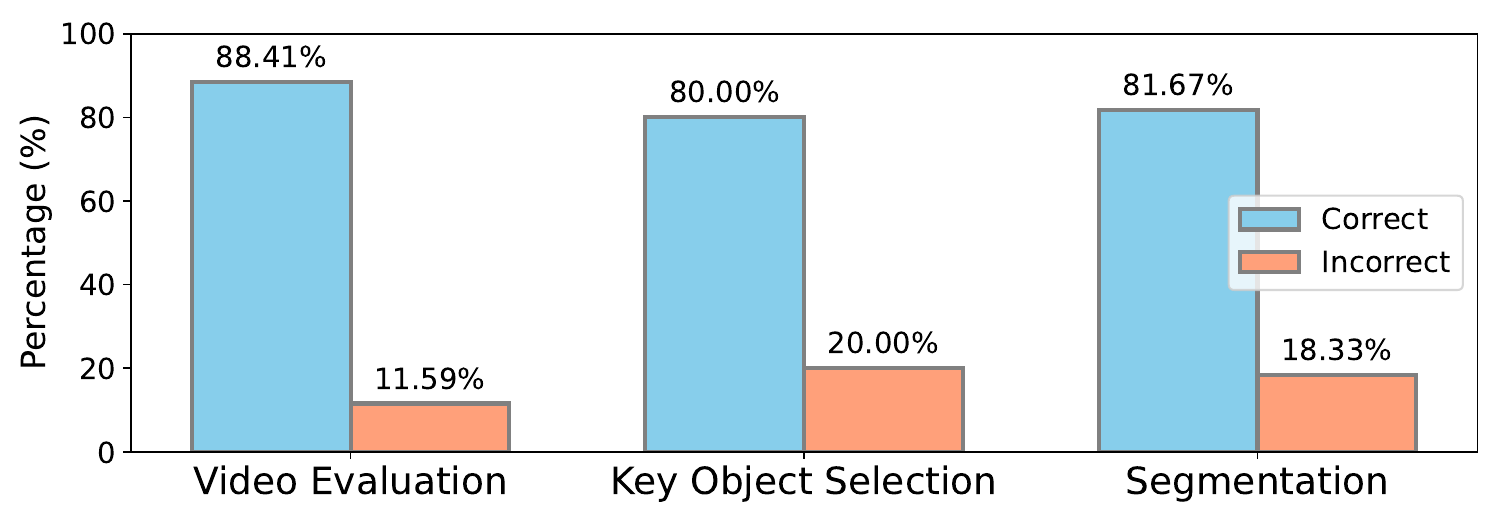}
    \caption{\textbf{Human error analysis results.}
    \ours{} consistently achieves approximately 80\% correctness across all components of the framework.
    }
    \label{fig:human_eval_results}
\end{figure}

For evaluation metrics, we mainly adopt the average text-video alignment score they proposed. 
Among their all text-video alignment scores (CLIP-Score, SD-Score, BLIP-BLEU, Detection-Score, Count-Score, Color-Score, Celebrity ID Score, and OCR-Score) we exclude Celebrity ID Score and OCR-Score since they are related to `face' and `text' categories. 
Therefore, we calculate the text-video alignment score as
Avg(CLIP-Score, SD-Score, BLIP-BLEU, DetectionScore, CountScore, ColorScore).
For overall video quality, we directly adopt their metrics including Inception Score~\citep{is} and Video Quality Assessment ($VQA_A$, $VQA_T$)~\citep{dover}. 
For the motion quality score, we calculate the weighted average score of the Action Recognition score (from VideoMAE~\citep{wang2023videomae}) and Average Flow score~\citep{teed2020raft} from the official EvalCrafter code.
For the temporal consistency score, we also calculate the weighted average score of Warping Error from optical flow~\citep{wang2023videomae} and CLIP-Temp~\citep{clip}. 
For the \textit{others} section of CogVideoX-5B, we report results on only 100 randomly sampled videos, as other baselines (e.g., OPT2I) require a significantly long refinement time (around 5h per one video refinement).

\paragraph{T2V-Compbench.}
Since \ours{} has strength in compositional generation, we adopt T2V-Compbench~\citep{sun2024t2v} and evaluate three dimensions: spatial relationships, generative numeracy, and consistent attribute binding. 
`Spatial relationships' requires the model to generate at least two objects while maintaining accurate spatial relationships (\eg{} `to the left of', `to the right of', `above', `below', `in front of') throughout the dynamic video. 
`Generative numeracy' specifies one or two object types, with quantities ranging from one to eight.
`Consistent attribute binding' contains color, shape, and texture attributes among two objects. 
Following \citep{sun2024t2v}, we adopt Video LLM-based metrics for consistent attribute binding and detection-based metrics for spatial Relationships and numeracy.

\section{Additional Quantitative Analysis}\label{sec:appendix-add-quantitative}

\subsection{Human Evaluation Results}\label{sec:appendix-error-analysis}
\begin{figure}[t]
  \centering
  \includegraphics[width=0.95\linewidth]{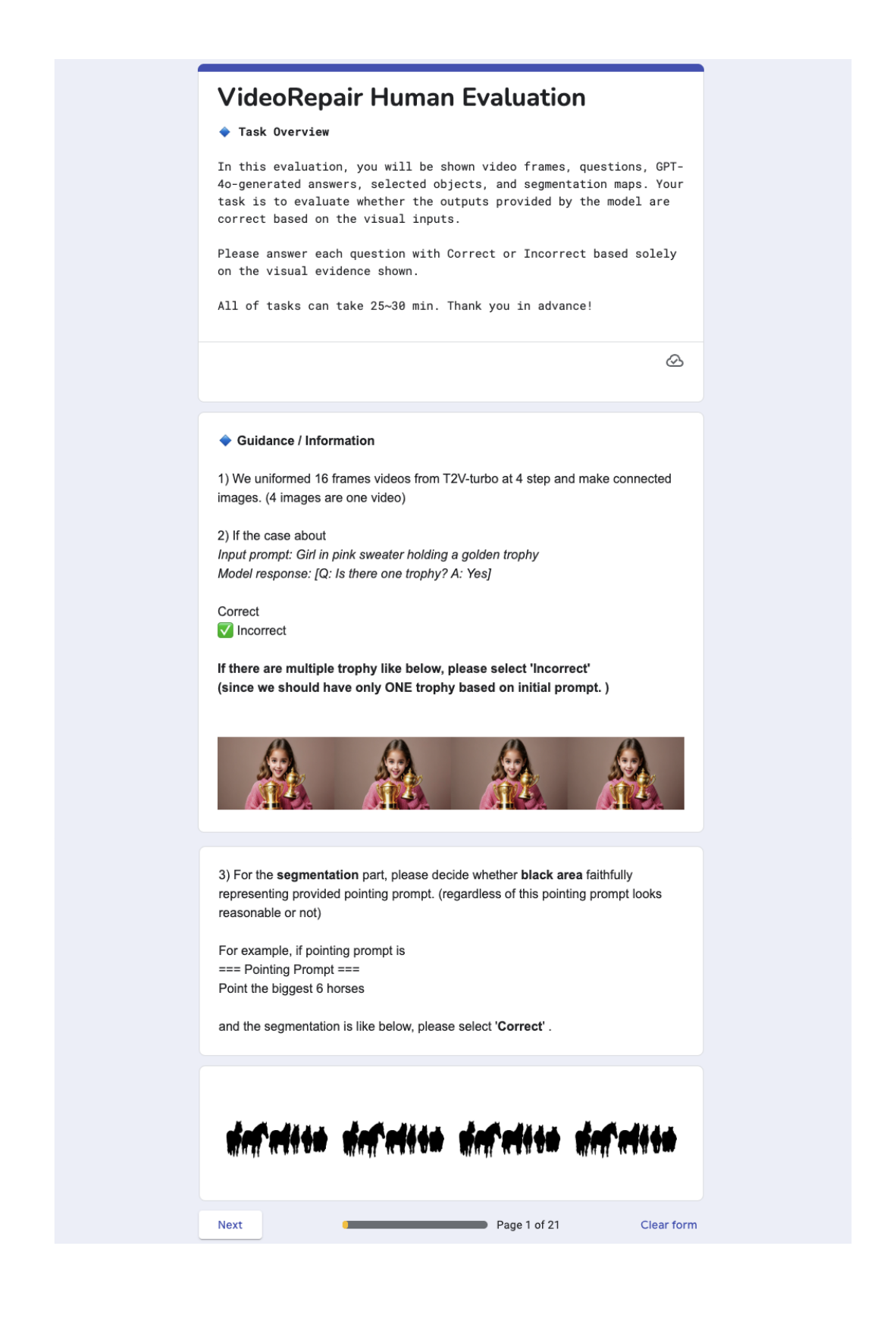}
   \caption{
   \textbf{A screenshot of questionnaires for error analysis.}
   }
   \label{fig:human_eval_screenshot1}
   \vspace{-0.1in}
\end{figure}

\begin{figure}[t]
  \centering
  \includegraphics[width=0.90\linewidth]{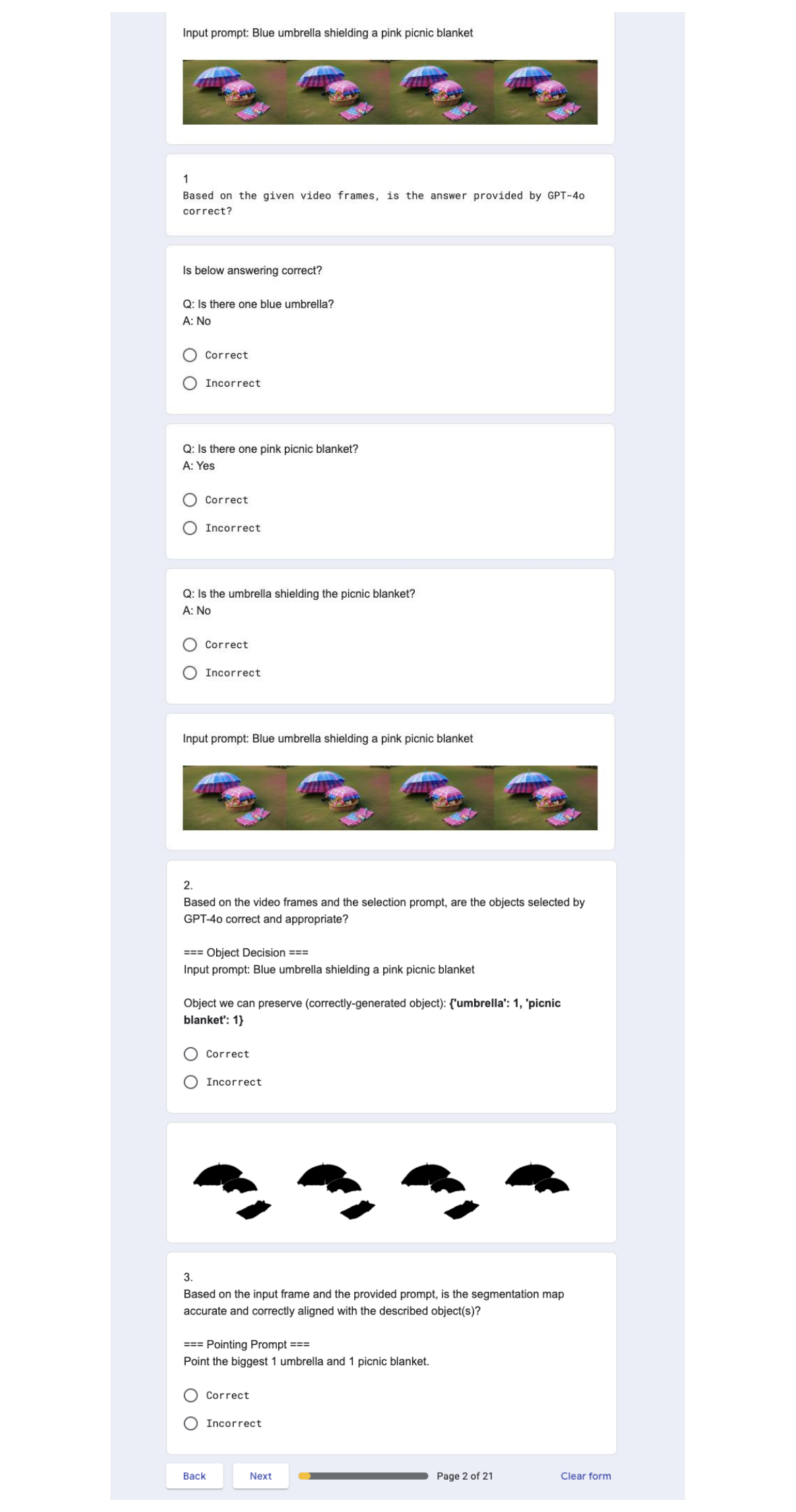}
   \caption{
   \textbf{A screenshot of questionnaires for error analysis.}
   }
   \label{fig:human_eval_screenshot2}
   \vspace{-0.1in}
\end{figure}

To analyze errors, we enlist three AI experts to assess the correctness of each component in \ours{}, including video evaluation, key object selection, and segmentation.
For video evaluation, annotators are provided with the initial text prompt and the corresponding video generated by T2V-Turbo, presented as a sequence of frames, along with MLLM-generated question–answer pairs. They are asked to judge whether the answers are correct. In particular, a generation is marked as \textit{incorrect} if the number of generated objects does not exactly match the count specified in the prompt. We include screenshots of the evaluation questionnaire and labeling instructions in \cref{fig:human_eval_screenshot1,fig:human_eval_screenshot2}.
For key object selection, we present the objects selected by GPT-4o and ask annotators to verify whether the identified objects $O^*$ and $N^*$ are correct.
For segmentation evaluation, annotators are given a pointing prompt (e.g., “Point to the largest umbrella and one picnic blanket”) together with the corresponding segmentation map, and are asked to assess whether the segmentation is well-aligned with the prompt.

Overall, human evaluation shows that \ours{} achieves approximately 80\% correctness across all components (\cref{fig:human_eval_results}). While errors remain at each stage, we attribute these limitations primarily to the backbone model and expect improvements with more advanced architectures. The inter-annotator agreement is 93.18\%, indicating strong consistency among annotators.

\begin{figure}[t]
    \centering
    \includegraphics[width=.75\linewidth]{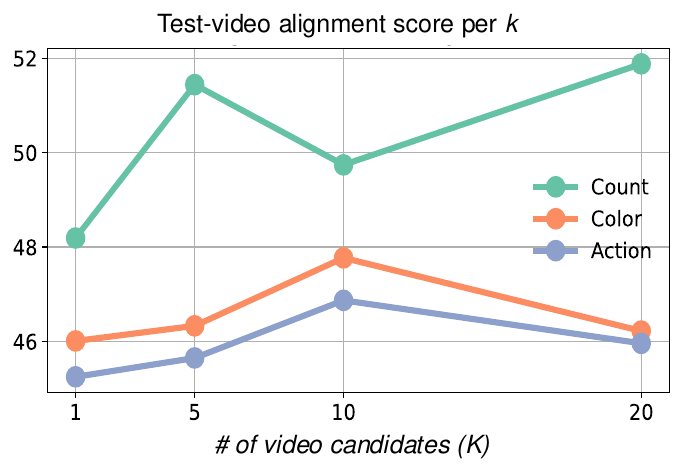}
    \caption{\textbf{Impact of the number of video candidates.}
    We vary the number of video candidates $K$ as 1, 5, 10, and 20 for ranking.}
    \label{fig:multi_k_plot}
\end{figure}

\begin{table*}[t]
\centering
\renewcommand{\arraystretch}{1.1}
\caption{\textbf{End-to-end inference latency (s).} We report end-to-end stage-wise runtime and total latency. }
\resizebox{\linewidth}{!}{%
\begin{tabular}{lcccccc}
\toprule
\textbf{Model} 
& \cellcolor[HTML]{E7F2DC}\textbf{Step 1 (Video Eval)} 
& \cellcolor[HTML]{FBF5DC}\textbf{Step 2 (Planning)} 
& \cellcolor[HTML]{FBE4E7}\textbf{Step 3 (Refinement)} 
& \cellcolor[HTML]{FBE4E7}\textbf{Step 3 (Ranking)} 
& \textbf{Total (s)} 
& \textbf{T2V Alignment} \\
\midrule
T2V-turbo & - & - & - & - & 3.55 & 43.54 \\
T2V-turbo (5 seed + random select) & - & - & - & - & 17.75 & 43.78 \\
T2V-turbo (5 seed + ranking select) & - & - & 17.75 & 24.2 & 41.95 & 44.68 \\
\midrule
\ours{} (k=1) & 5.33 & 22.1 & 5.35 & - & 32.78 & 46.53 \\
\ours{} (k=5) -- w/ BLIP ranking & 5.21 & 22.5 & 24.8 & - & 52.51 & 47.77 \\
\ours{} (k=5) -- w/ DSG ranking & 5.21 & 22.5 & 24.8 & 23.5 & 76.01 & \textbf{47.91} \\
\bottomrule
\end{tabular}
}
\label{tab:inference-latency}
\end{table*}

\begin{table*}[t]
\centering
\renewcommand{\arraystretch}{1.1}
\caption{\textbf{Cost comparison with other baselines.} We report the number of LLM/MLLM calls, external model calls (e.g., detection and segmentation), and stage-wise runtime.}
\resizebox{\linewidth}{!}{%
\begin{tabular}{lcccccc}
\toprule
\textbf{Model} 
& \textbf{\# LLM/MLLM Calls} 
& \textbf{\# External Model Calls} 
& \textbf{Detection+Planning} 
& \textbf{Refinement+Ranking} 
& \textbf{Total Latency} 
& \textbf{T2V Alignment} \\
\midrule
T2V-turbo & - & - & - & - & 3.55 & 43.54 \\
LLM Paraphrasing (k=5) & 5 & - & 1.3 & 17.7 & 19.0 & 44.6 \\
OPT2I (k=5) & 61 & - & 75.3 & 213 & 288.3 & 45.6 \\
SLD & 17 & 16 & 166.7 & 378.5 & 545.2 & 44.5 \\
\ours{} (k=5, BLIP ranking) & 6 & 4 & 27.7 & 24.8 & 52.5 & \textbf{47.7} \\
\bottomrule
\end{tabular}
}
\label{tab:cost}
\end{table*}

\subsection{Cost and Efficiency Analysis}\label{sec:appendix-inference-latency}

\textbf{ End-to-end inference latency.} To provide a complete end-to-end cost analysis, we report full wall-clock latency, per-stage breakdown, and compute-matched baselines (seed generation) in \cref{tab:inference-latency}. All experiments are conducted on two NVIDIA RTX 6000 40GB GPUs, and running results are averaged over the EvalCrafter subsets (count, color, and action). 
 Compute-matched sampling baselines (e.g., 5-seed generation with ranking) improve only marginally (44.68), indicating that our gains are not attributable to increased sampling alone.

\textbf{Runtime and model call.} \cref{tab:cost} provides a cost comparison with refinement baselines, including the \# of LLM/MLLM calls, external model calls (e.g., segmentation and detection), and stage-wise runtime (s). For fair comparison, we match the candidate count across methods.
The results show that while OPT2I and SLD incur substantial detection/planning and refinement overhead (288–545s total latency), \ours{} achieves higher T2V alignment (47.77) with significantly lower latency (52.51s), yielding a \textbf{5–10× speedup} over these refinement-based baselines.

\subsection{Scaling Behavior}\label{sec:appendix-scaling}
To further characterize scaling behavior, we analyze the impact of (i) number of candidates ($K$), (ii) video length (16 v.s. 81 frames), and (iii) video resolution (480x832 v.s. 720x1280). All experiments are conducted using T2V-turbo and Wan2.1, evaluated on two NVIDIA H100 80GB GPUs.

\textbf{Increasing \# of Video Candidates.} 
To evaluate the impact of video ranking, we vary the number of video candidates as $K=1, 5, 10, \text{and~} 20$ during the ranking process.
The variation among video candidates arises from different random seeds used to initialize $\epsilon_0'$.
For example, video ranking is not applied when $K=1$, and only one refinement is produced using a single random seed noise $\epsilon_0'$.
For ranking metrics, we rely on the video score across all ablation studies.
As depicted in \cref{fig:multi_k_plot}, higher $K$ values (5, 10, and 20) consistently yield higher scores across all categories than $K=1$. This trend is particularly prominent in the `count' category, where increasing $K$ leads to noticeable performance improvements, highlighting the importance of considering multiple candidates for ranking.

\textbf{Increasing Video Length and Resolution.} 
As shown in \cref{tab:resolution-scaling,tab:length-scaling}, increasing video length and resolution leads to longer refinement time. This is expected because localized refinement performs mask-based noise re-initialization and runs separate diffusion paths for preserved and refined regions, followed by fusion and candidate ranking, causing the cost to scale with video length, spatial refinement area, and the number of seeds. 
Importantly, this cost profile reflects an implementation characteristic rather than a fundamental limitation. The overhead can be mitigated through temporal sparsification (refining only key frames), spatial compression (operating on downsampled or region-focused representations), and acceleration strategies such as lightweight refinement backbones or feature caching.

\begin{table}[t]
\centering
\small
\caption{\textbf{Scaling with respect to video length.} Increasing the number of frames incurs a higher computational cost due to longer diffusion trajectories and expanded spatio-temporal processing.}
\renewcommand{\arraystretch}{1.1}
\resizebox{\linewidth}{!}{%
\begin{tabular}{lccccc}
\toprule
\textbf{Model} & \cellcolor[HTML]{E7F2DC}\textbf{Step1} & \cellcolor[HTML]{FBF5DC}\textbf{Step2} & \cellcolor[HTML]{FBE4E7}\textbf{Step3} & \cellcolor[HTML]{FBE4E7}\textbf{Ranking} & \textbf{Total (s)} \\
\midrule
Wan2.1 (16f) & - & - & - & - & 19.9 \\
Wan2.1 (16f) + Ours (k=1) & 6.21 & 22.3 & 37.2 & - & 65.7 \\
Wan2.1 (16f) + Ours (k=5) & 6.21 & 22.3 & 185 & 21.3 & 234.8 \\
\midrule
Wan2.1 (81f) & - & - & - & - & 143.1 \\
Wan2.1 (81f) + Ours (k=1) & 6.3 & 50.2 & 279.6 & - & 336.1 \\
Wan2.1 (81f) + Ours (k=5) & 6.3 & 50.3 & 1300 & 20.5 & 1377.1 \\
\bottomrule
\end{tabular}
}
\label{tab:length-scaling}
\end{table}

\begin{table}[t]
\centering
\renewcommand{\arraystretch}{1.1}
\caption{\textbf{Scaling with respect to resolution.} Higher spatial resolution increases computational cost, as localized diffusion operates over larger spatial regions.}
\resizebox{\linewidth}{!}{%
\begin{tabular}{lccccc}
\toprule
\textbf{Model} & \cellcolor[HTML]{E7F2DC}\textbf{Step1} & \cellcolor[HTML]{FBF5DC}\textbf{Step2} & \cellcolor[HTML]{FBE4E7}\textbf{Step3} & \cellcolor[HTML]{FBE4E7}\textbf{Ranking} & \textbf{Total (s)} \\
\midrule
Wan2.1 (480$\times$832) & - & - & - & - & 143.1 \\
Wan2.1 (480$\times$832) + Ours (k=1) & 6.3 & 50.2 & 279.6 & - & 336.1 \\
Wan2.1 (480$\times$832) + Ours (k=5) & 6.3 & 50.3 & 1300 & 20.5 & 1377.1 \\
\midrule
Wan2.1 (720$\times$1280) & - & - & - & - & 596.9 \\
Wan2.1 (720$\times$1280) + Ours (k=1) & 6.1 & 51.4 & 1148.2 & - & 1205.7 \\
Wan2.1 (720$\times$1280) + Ours (k=5) & 6.2 & 51.6 & 5743.1 & 21.5 & 5822.4 \\
\bottomrule
\end{tabular}
}
\label{tab:resolution-scaling}
\end{table}

\begin{figure*}[h]
    \centering
    \includegraphics[width=1.0\linewidth]{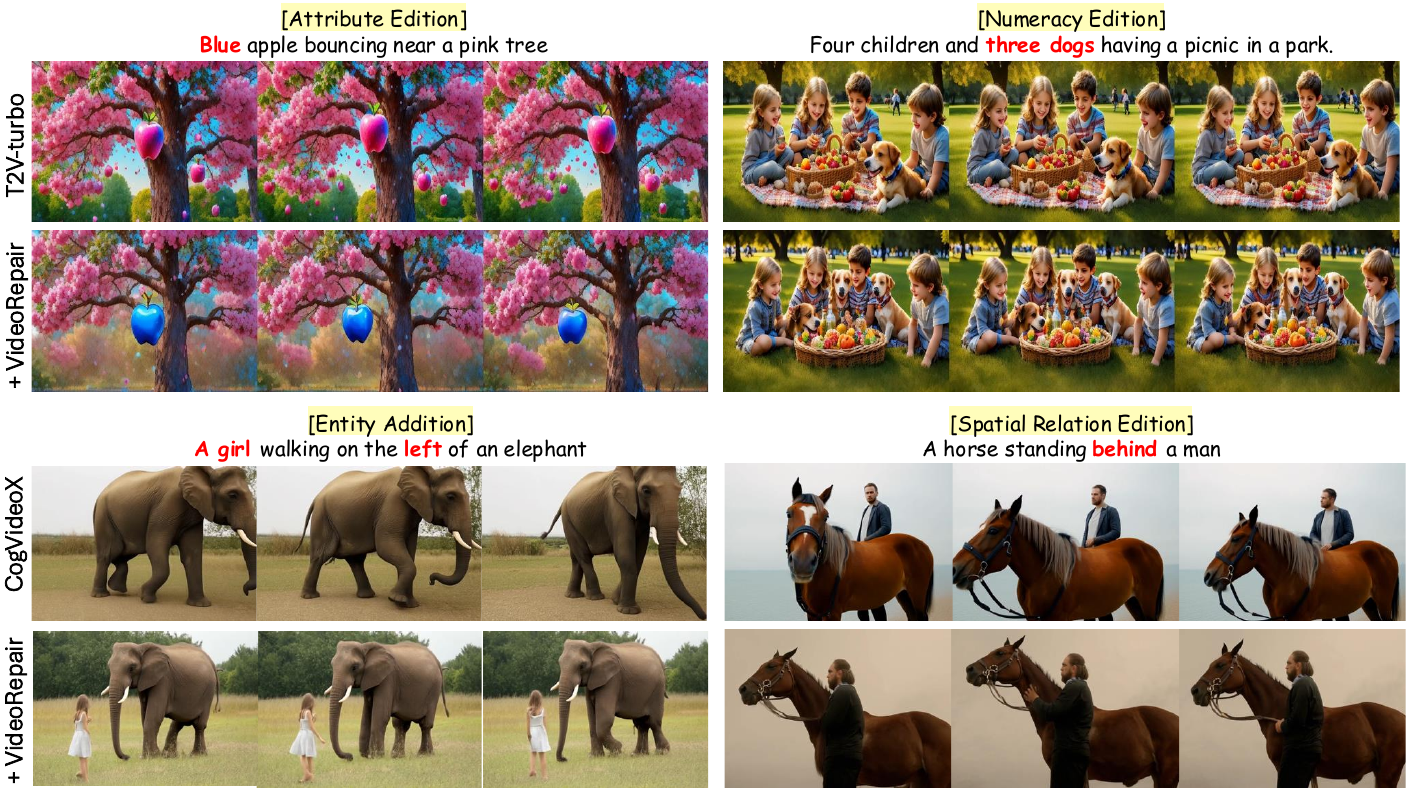}
    \caption{\textbf{Scalability of \ours{}.} 
    We illustrate the scalability of \ours{} including attribute, numeracy, spatial relation edition, and entity addition from T2V-turbo and CogVideoX backbone. 
    }
    \label{fig:qual_scale_category_examples}
\end{figure*}

\subsection{Global Refinement Results}\label{sec:appendix-global-refinement}
\Cref{tab:global_refine} reports results on the \textit{Other} section of EvalCrafter, 
which includes camera movement, landscape, and style prompts. 
Applying \ours{} yields consistent improvements across all three categories, 
with gains of +1.21 in camera movement, +1.77 in landscape, and +0.93 in style. 
These results highlight that \ours{} not only enhances core compositional attributes (e.g., count, color, action) but also extends effectively to broader aspects of video quality such as dynamics, scenery, and artistic style.

\begin{table}[t]
\centering
\caption{\textbf{\ours{} performance on the EvalCrafter \textit{`Other'} section.} 
\ours{} consistently improves video quality in camera movement, landscape, and style categories over the initial generations.}
\resizebox{.99\linewidth}{!}{
\begin{tabular}{lccc}
\toprule
 & Camera movement & Landscape & Style \\
\midrule
Initial video (T2V-turbo) & 44.02 & 48.94 & 42.70 \\
+ \ours{} & \textbf{45.23} & \textbf{50.71} & \textbf{43.63} \\
\bottomrule
\end{tabular}
}
\label{tab:global_refine}
\end{table}

\subsection{Full Results on EvalCrafter}\label{sec:appendix-evalcrafter-full}
To enable direct comparison with prior work that reports the official average across all splits, we evaluated \ours{} on the full EvalCrafter~\cite{liu2023evalcrafter} benchmark with T2V-turbo, including previously excluded categories (text, camera motion, and face).
As shown in \cref{tab:evalcrafter-full}, \ours{} consistently improves T2V alignment on the full benchmark while maintaining comparable visual quality, motion quality, and temporal consistency. 
The gains in motion quality and temporal consistency are intentionally modest, as \ours{} is designed to correct semantic misalignment while preserving the original video dynamics rather than re-synthesizing motion. Overall, the results confirm that the improvements are not limited to selected splits but generalize to the complete EvalCrafter evaluation protocol while maintaining stable video quality metrics.

\begin{table}[t]
\centering
\caption{\textbf{Evaluation on full EvalCrafter benchmark.} For direct comparison with prior work, we additionally report the official average across all splits.}
\resizebox{\linewidth}{!}{%
\begin{tabular}{lccccc}
\toprule
\textbf{Model} 
& \makecell{\textbf{Visual} \\ \textbf{Quality}} 
& \makecell{\textbf{T2V} \\ \textbf{Alignment}} 
& \makecell{\textbf{Motion} \\ \textbf{Quality}} 
& \makecell{\textbf{Temporal} \\ \textbf{Consistency}} 
& \makecell{\textbf{Final} \\ \textbf{Score}} \\
\midrule
Zeroscope     & 53.41 & 51.21 & 53.61 & 58.91 & 217.14 \\
MoonValley    & 69.53 & 50.66 & 55.46 & 65.25 & 240.90 \\
Show-1        & 52.19 & 62.07 & 53.74 & 60.83 & 228.83 \\
T2V-turbo     & 61.23 & 59.87 & 55.94 & 62.78 & 239.82 \\
+ \ours{} & \textbf{61.26} & \textbf{62.30} & \textbf{55.90} & \textbf{62.88} & \textbf{242.34} \\
\bottomrule
\end{tabular}
}
\label{tab:evalcrafter-full}
\end{table}

\section{Additional Qualitative Results}\label{sec:appendix-additional-qual}

\subsection{Iterative Refinement}\label{sec:appendix-iter-refinement}
\begin{figure}[t]
    \centering
    \includegraphics[width=1.0\linewidth]{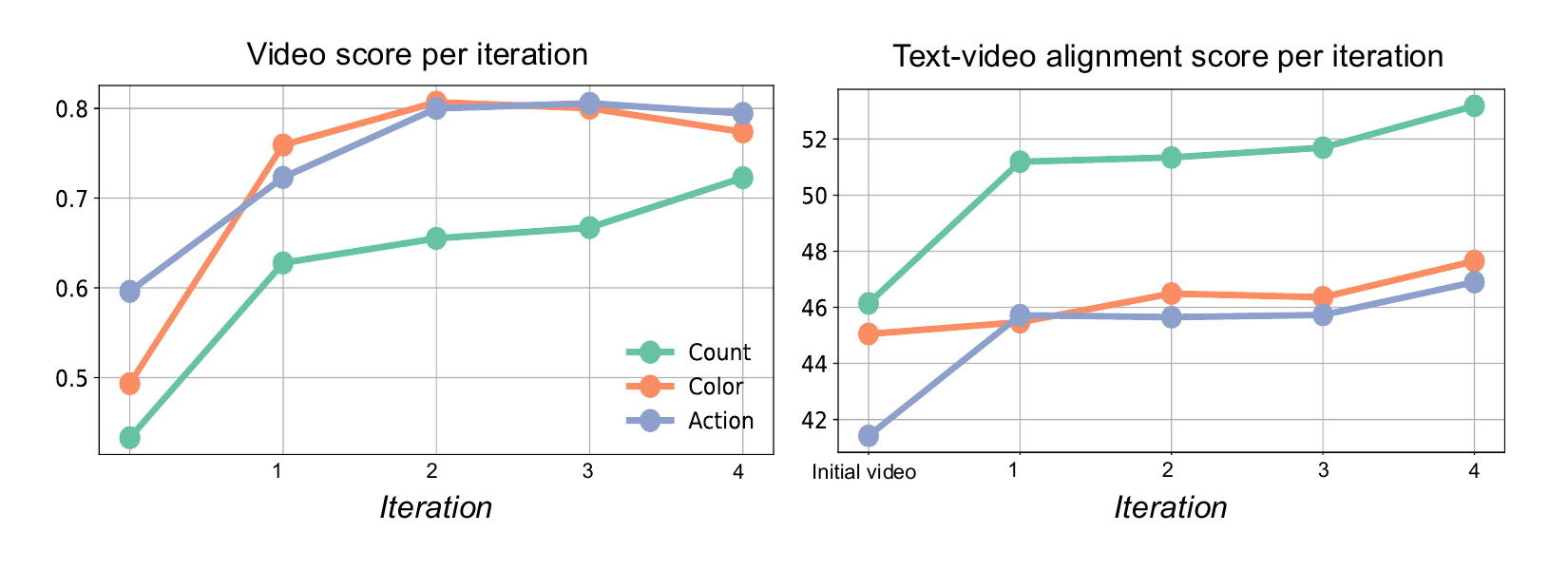}
    \caption{\textbf{Impact of iterative refinement.}
    Iterative refinement gradually improves the video score and text-video alignment score on all three prompt categories (count/color/action) of EvalCrafter.
    }
    \label{fig:multi_round_plot}
\end{figure}

We observe the effect of applying \ours{} iteratively to further improve text–video alignment. At each step, we monitor the video score and terminate the refinement process once it reaches the maximum value of 1.0. We use video ranking with $K=5$.
As shown in \cref{fig:multi_round_plot}, iterative refinement consistently improves performance across all three splits (count, color, and action) in EvalCrafter. 

\begin{figure*}[t]
    \centering
    \includegraphics[width=1.0\linewidth]{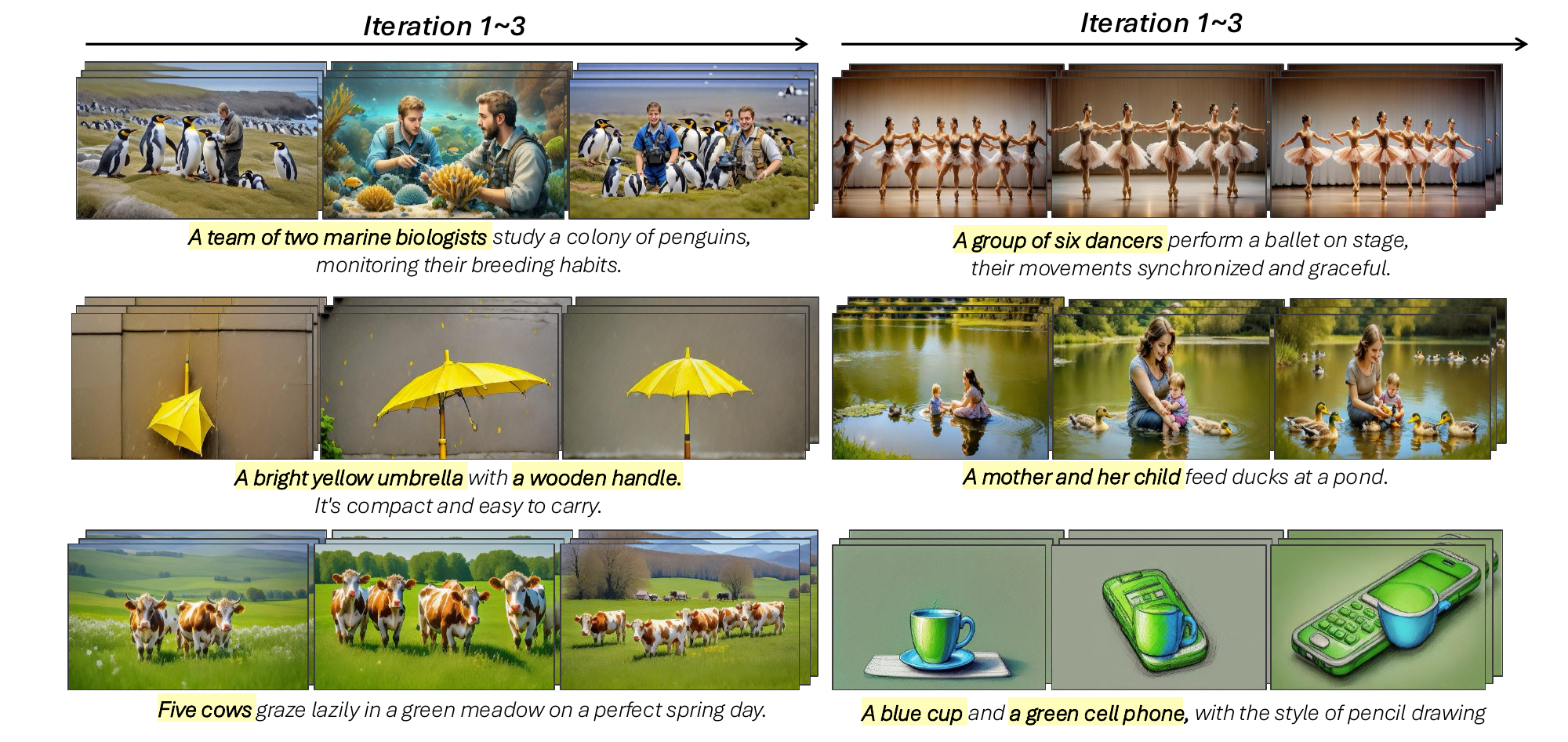}
    \caption{\textbf{Videos generated using iterative refinement with \ours{}.} We depict iterative refinement results generated from T2V-Turbo. Overall, \ours{} progressively enhances text-video alignment with each refinement step. }
    \label{fig:fig_iterative}
\end{figure*}

Qualitative examples in \cref{fig:fig_iterative} further illustrate this trend, where we compare the initial video with the first and second refinement results generated by T2V-Turbo. Overall, \ours{} progressively enhances text–video alignment over successive refinement steps.
In numeracy-related cases (\eg{}, six dancers, five cows), \ours{} iteratively adjusts object counts by adding or removing instances to match the prompt. When objects are missing (\eg{}, biologists, ducks), the model successfully introduces additional instances while preserving the original scene context. For attribute-related prompts (\eg{}, yellow umbrella, blue cup), \ours{} refines object attributes, such as introducing a wooden handle to the umbrella and strengthening the blue color of the cup. These results demonstrate that \ours{} effectively improves both object count and attribute alignment in a progressive and high-fidelity manner.

\subsection{Object Selection}\label{sec:appendix-obj-selection}
\begin{figure}[t]
  \centering
  \includegraphics[width=0.99\linewidth]{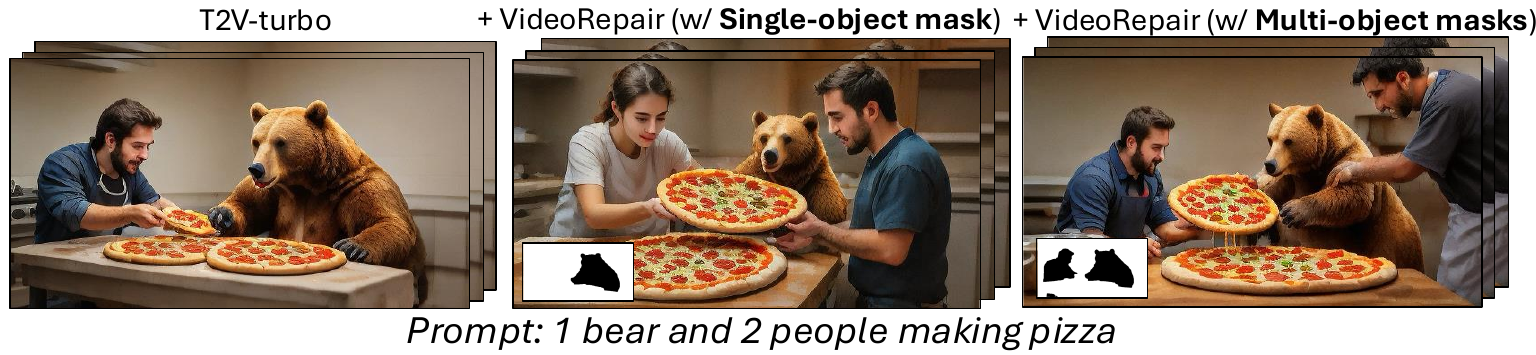}
   \caption{
   \textbf{Single-object mask vs. Multi-object mask.}
   }
   \label{fig:multi_can}
   \vspace{-0.1in}
\end{figure}

In refinement planning, we select the largest candidate among the correct objects.
This approach can be seamlessly extended to select multiple correct objects when the number of objects in the initial video ($n^{o^*}_v$) meets or exceeds the prompt's specification ($n^{o^*}_p$).
We implement this extension to enable the formulation of \textbf{object-wise pointing prompts} and the generation of multiple masks to preserve these objects. As shown in \cref{fig:multi_can}, this version can preserve \textit{a bear} and \textit{a man} while automatically refining the video to add \textit{an additional person}.

\subsection{Moving Key Objects.}
\begin{figure}[t]
  \centering    \includegraphics[width=.99\linewidth]{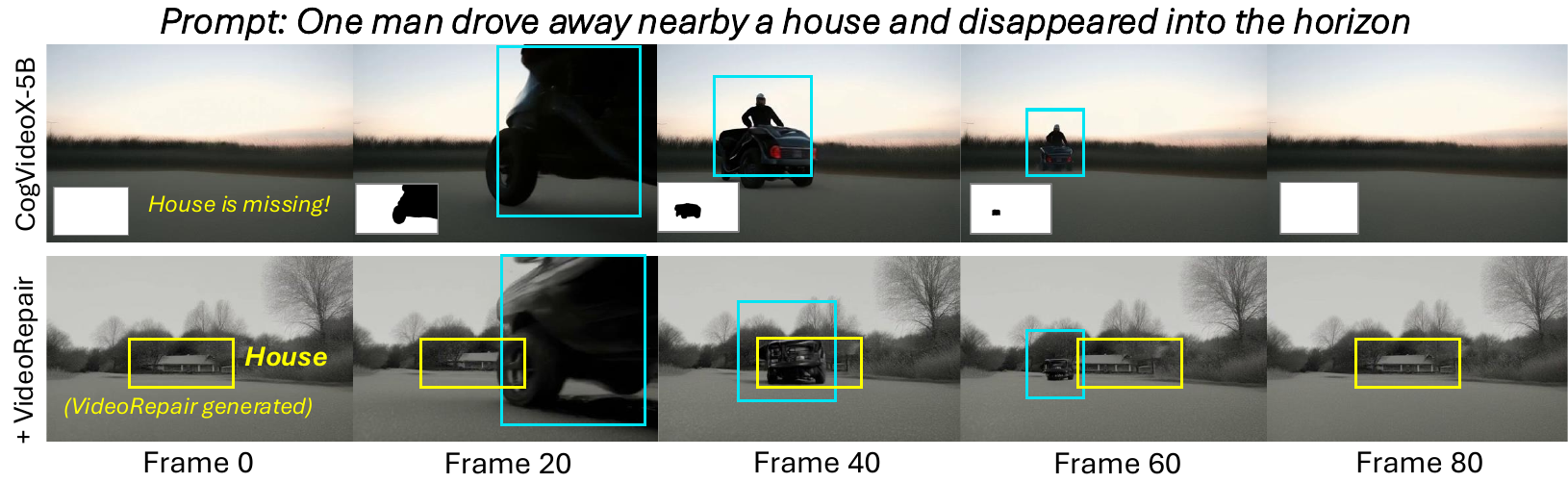}
   \caption{
   \textbf{Refining videos when the key object disappears.}
   \ours{} successfully preserves disappearing objects (\textit{car}) while incorporating previously missed objects (\textit{house}).
   }
   \label{fig:moving_object}
\end{figure}

In long videos (e.g., CogVideoX-generated videos with 81 frames), key objects may disappear or newly appear across different frames. 
As shown in \cref{fig:moving_object}, \ours{} effectively captures moving key objects $O^*$ using frame-wise masks $M$. This example illustrates how frame-wise masks help handle changes in object count and attributes - preserving disappearing objects (\textit{car}) while incorporating previously missed objects (\textit{house}).

\subsection{Step-by-Step Illustration}\label{sec:appendix-step-illustration}

\begin{figure*}[h]
    \centering
    \includegraphics[width=1.0\linewidth]{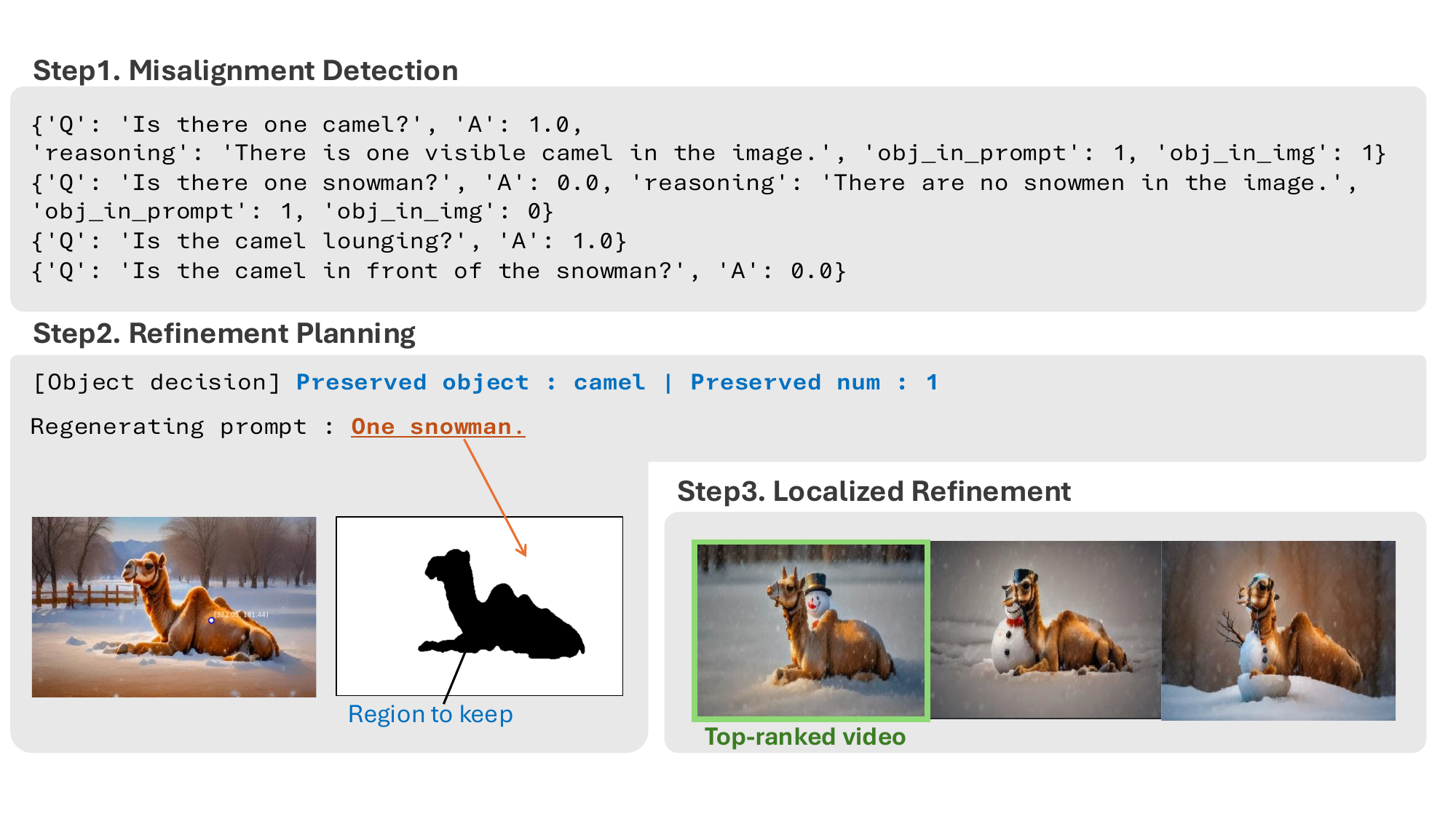}
    \caption{\textbf{Output from each step of \ours{}.} We illustrate whole outputs from each step of \ours{}.}
    \label{fig:visual_example1}
\end{figure*}

\begin{figure*}[h]
    \centering
    \includegraphics[width=1.0\linewidth]{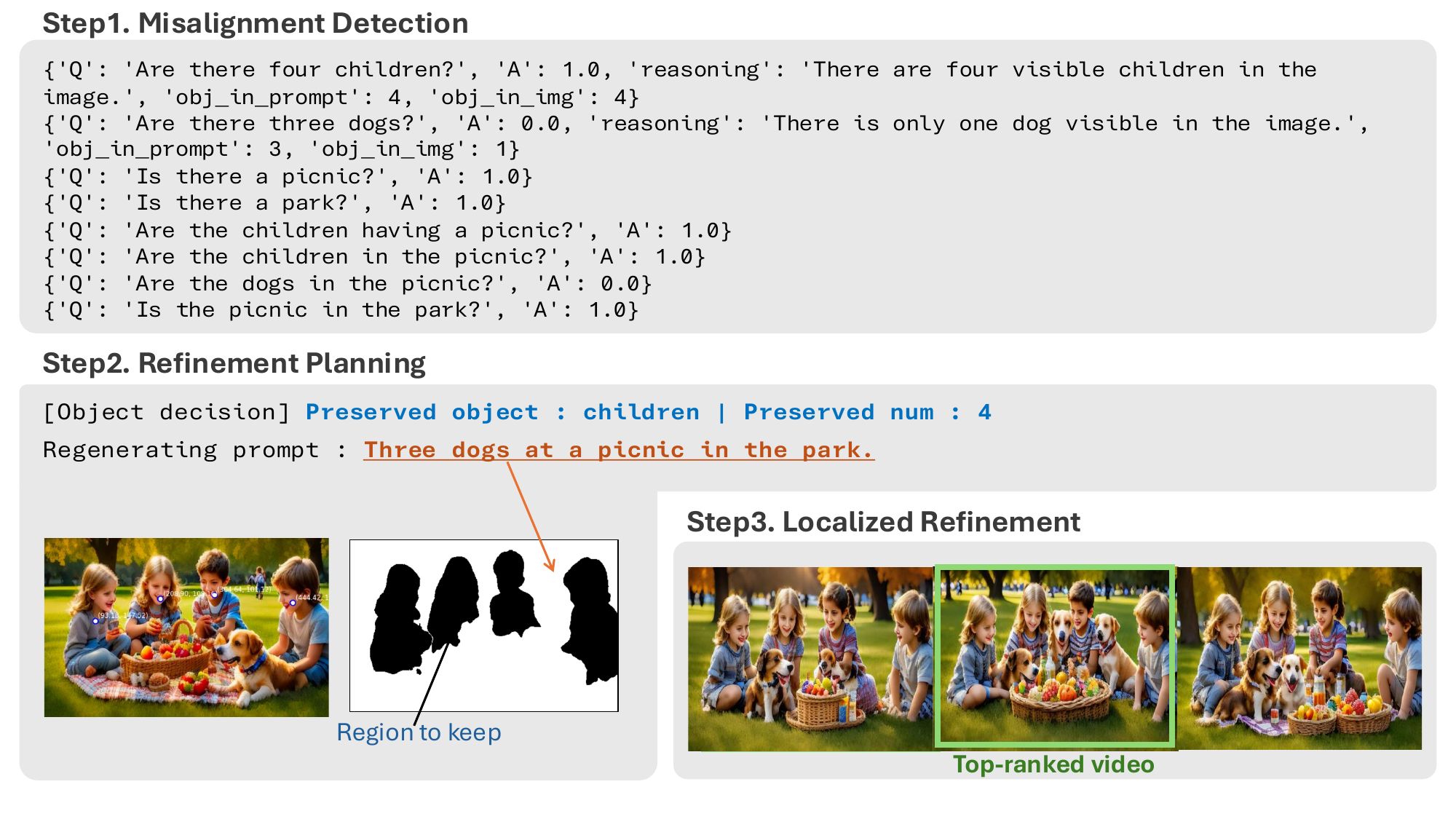}
    \caption{\textbf{Output from each step of \ours{}.} We illustrate whole outputs from each step of \ours{}.}
    \label{fig:visual_example2}
\end{figure*}

In \cref{fig:visual_example1,fig:visual_example2}, we provide detailed illustrations of all three \ours{} steps.

\subsection{Comparison with Baselines}\label{sec:appendix-comparison-baseline}
\begin{figure*}[h]
    \centering
    \includegraphics[width=1.0\linewidth]{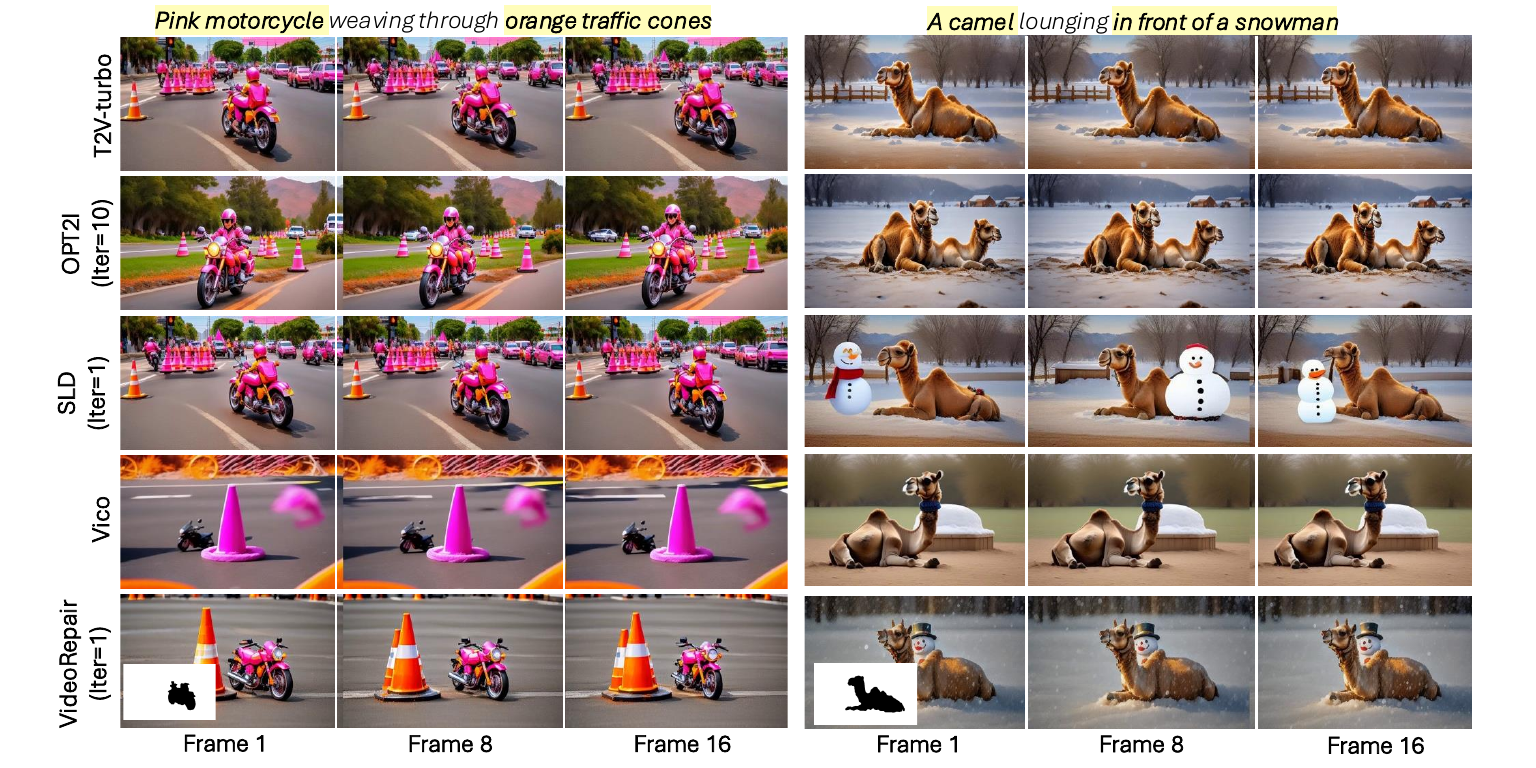}
    \caption{\textbf{Qualitative examples from T2V-turbo.} }
    \label{fig:qual_turbo1}
\end{figure*}

\begin{figure*}[h]
    \centering
    \includegraphics[width=1.0\linewidth]{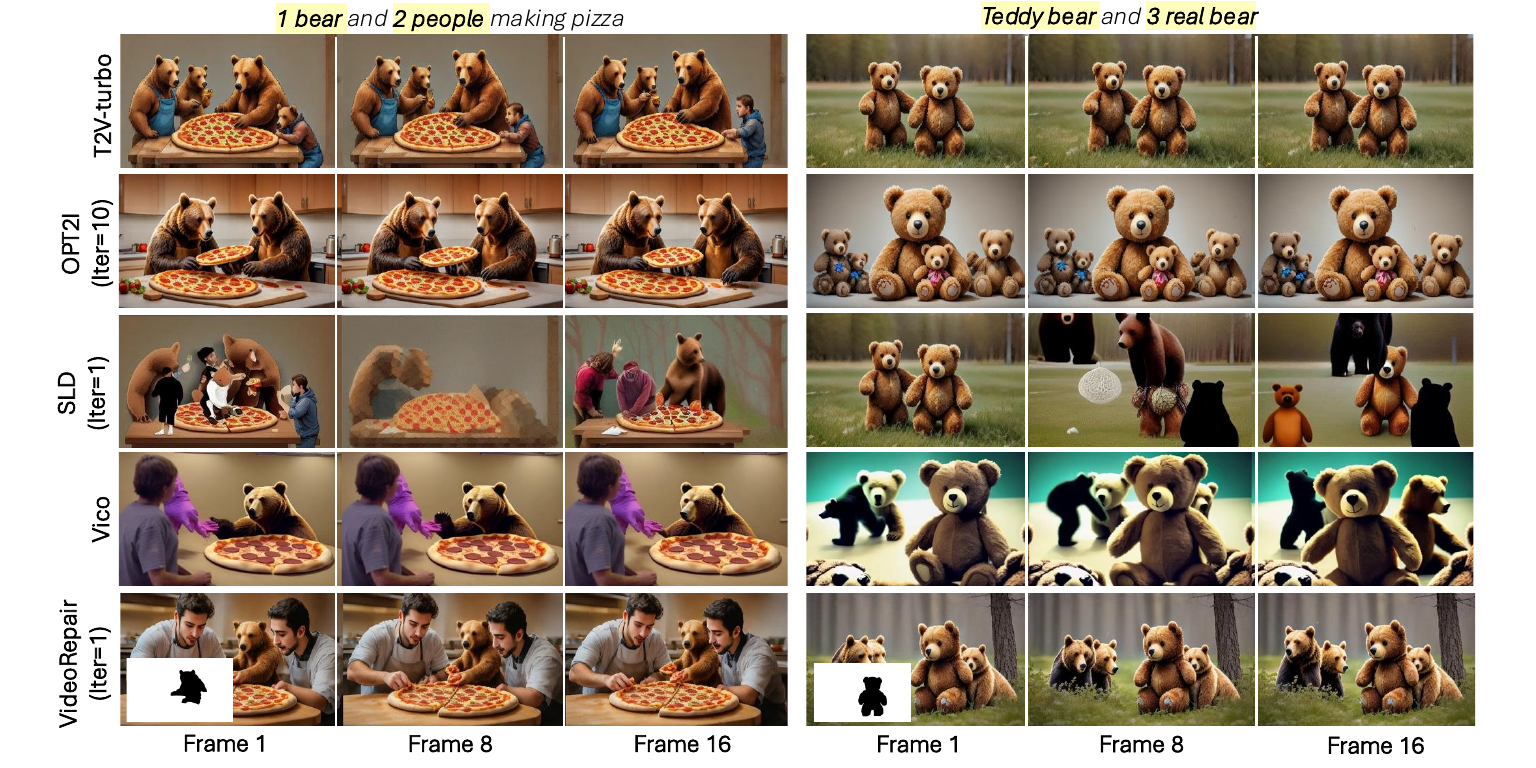}
    \caption{\textbf{Qualitative examples from T2V-turbo.} }
    \label{fig:qual_turbo2}
\end{figure*}

\begin{figure*}[h]
    \centering
    \includegraphics[width=1.0\linewidth]{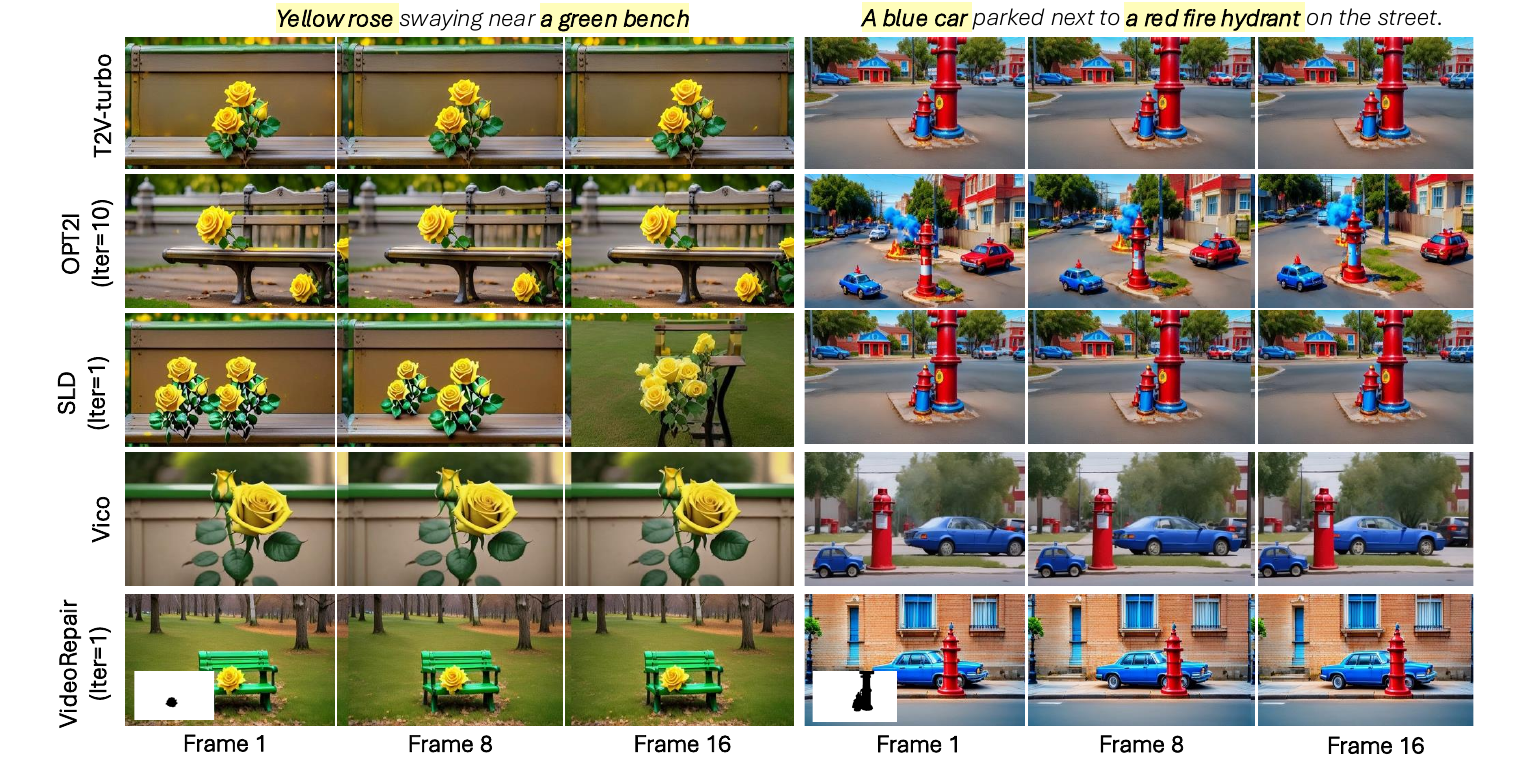}
    \caption{\textbf{Qualitative examples from T2V-turbo.} }
    \label{fig:qual_turbo3}
\end{figure*}

\begin{figure*}[h]
    \centering
    \includegraphics[width=1.0\linewidth]{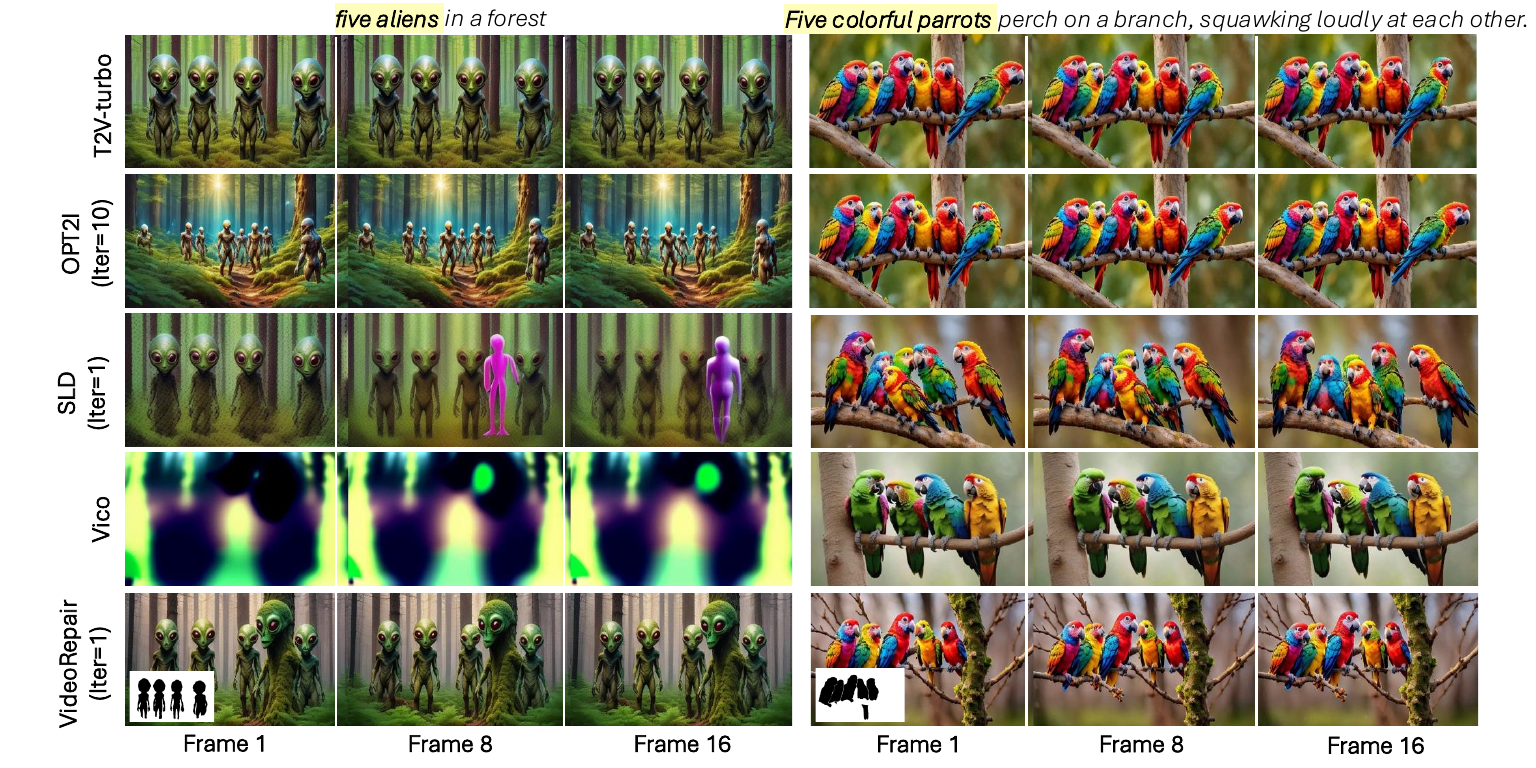}
    \caption{\textbf{Qualitative examples from T2V-turbo.} }
    \label{fig:qual_turbo4}
\end{figure*}

\begin{figure*}[h]
    \centering
    \includegraphics[width=1.0\linewidth]{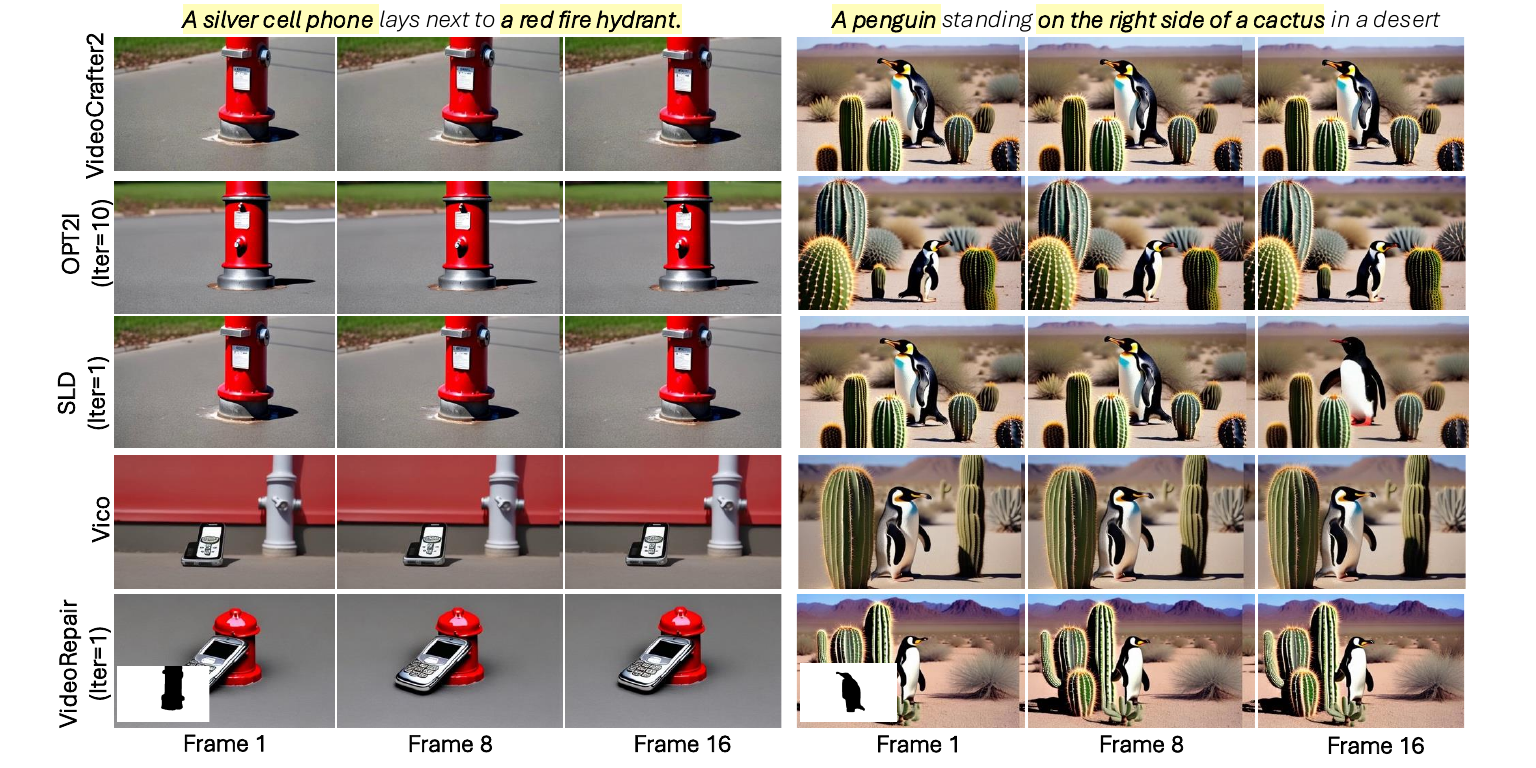}
    \caption{\textbf{Qualitative examples from VideoCrafter2.} }
    \label{fig:qual_vc1}
\end{figure*}

\begin{figure*}[h]
    \centering
    \includegraphics[width=1.0\linewidth]{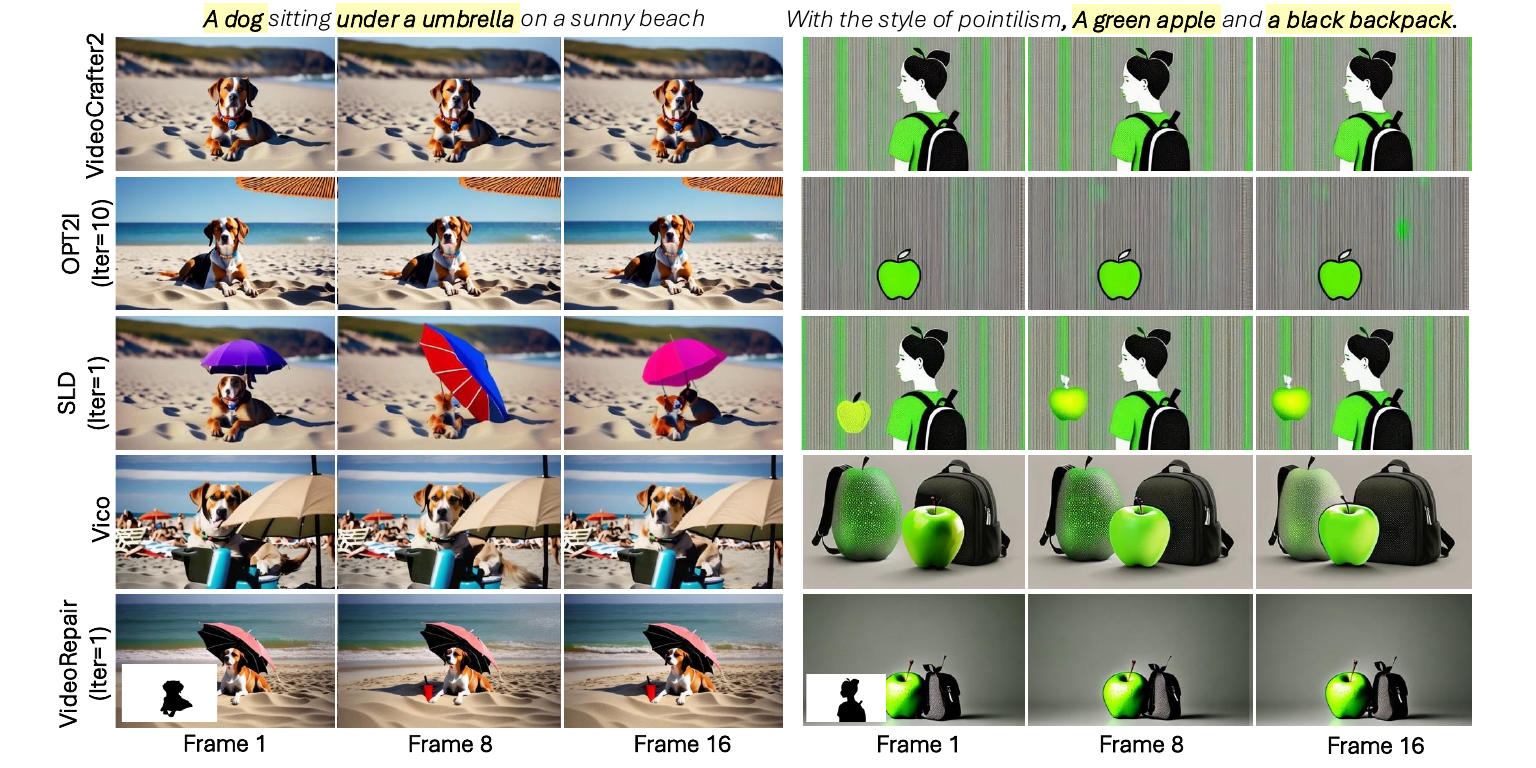}
    \caption{\textbf{Qualitative examples from VideoCrafter2.} }
    \label{fig:qual_vc2}
\end{figure*}

\begin{figure*}[h]
    \centering
    \includegraphics[width=1.0\linewidth]{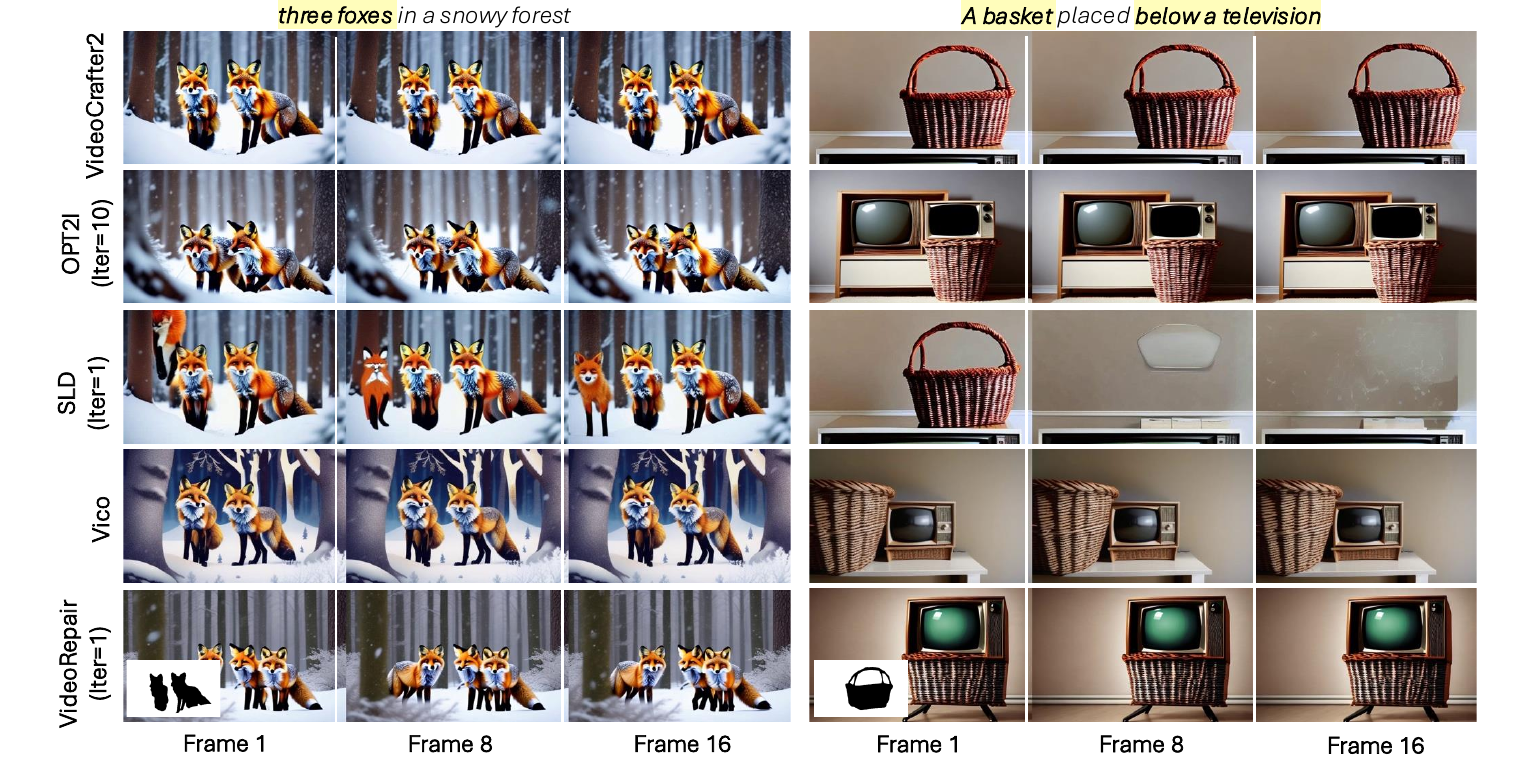}
    \caption{\textbf{Qualitative examples from VideoCrafter2.} }
    \label{fig:qual_vc3}
\end{figure*}

We present additional qualitative comparisons with baseline methods (OPT2I~\citep{opt2i}, SLD~\citep{sld}, and Vico~\citep{vico}) in \cref{fig:qual_turbo1,fig:qual_turbo2,fig:qual_turbo3,fig:qual_turbo4,fig:qual_vc1,fig:qual_vc2,fig:qual_vc3}. 
These examples address a variety of failure cases commonly observed in T2V models, including inaccuracies in object count and attribute depiction, as highlighted in our main paper.
\cref{fig:qual_turbo1,fig:qual_turbo2,fig:qual_turbo3,fig:qual_turbo4} correspond to results from T2V-Turbo, while \cref{fig:qual_vc1,fig:qual_vc2,fig:qual_vc3} showcase examples from VideoCrafter2. Additionally, we provide binary segmentation masks that identify preserved areas (in black) and updated areas (in white).

Across these examples, \ours{} effectively preserves the $O^*$ areas while refining the remaining regions using $p^r$. For instance, in \cref{fig:qual_turbo1}, the camel from the original T2V-Turbo video is preserved, and a snowman is successfully added. In contrast, while SLD also leverages DDIM inversion to preserve objects, it often fails to integrate new objects seamlessly.
We also visualize the scalability of \ours{} in \cref{fig:qual_scale_category_examples}.

\section{Future Work}\label{sec:appendix-future-work}
To further address the identified failure modes on \cref{tab:failure-breakdown}, we plan to incorporate targeted mitigation strategies tailored to each category. For instance, we can introduce confidence-based gating mechanisms to reduce QA hallucination by skipping low-confidence refinements and enforcing stricter acceptance criteria, with a fallback to global generation when the video score is unreliable. To alleviate mask drift, temporal mask smoothing can be applied to improve cross-frame consistency. Boundary artifacts can be mitigated by replacing hard binary masks with soft blending masks (e.g., Gaussian-blurred masks) for smoother transitions. Additionally, identity and motion inconsistencies can be reduced by increasing preservation weights in the joint optimization objective and adopting soft transition masks to better anchor preserved content. Importantly, as \ours{} is a training-free and model-agnostic framework, these improvements can be seamlessly integrated without retraining, enabling flexible extensions and systematic evaluation in future work.


\newpage

\newpage

\newpage

